\documentclass[11pt,a4paper]{ouparticle}

\usepackage{multicol}
\usepackage{multirow}
\usepackage[numbers,sort&compress]{natbib}
\usepackage[labelfont=bf]{caption}
\usepackage{pdfpages}

\newcommand{\data}{\mathbf{D}}
\newcommand{\alignment}[1]{\mathcal{A}_{#1}}
\newcommand{\seqS}[1]{S_{#1}}
\newcommand{\seqT}[1]{T_{#1}}
\newcommand{\match}{\texttt{match}}
\newcommand{\insertion}{\texttt{insert}}
\newcommand{\delete}{\texttt{delete}}
\newcommand{\seqpair}[1]{\left< \seqS{#1},\seqT{#1} \right>}
\newcommand{\alseqpair}[1]{\left< \alignment{#1},\seqS{#1},\seqT{#1} \right> }

\newcommand{\submat}[1]{\mathbf{M}_{#1}}
\newcommand{\nullmodel}{\mathbf{P}}
\newcommand{\dirparamst}[1]{\boldsymbol{\alpha}(#1)}
\newcommand{\dirparams}{\boldsymbol{\alpha}}
\newcommand{\stateParams}{\mathbf{\Theta}}
\newcommand{\set}[1]{\{ #1 \}}
\newcommand{\conditionalmat}{\mathbf{C}}

\newcommand{\prob}[1]{\text{Pr($#1$)}}
\newcommand{\I}[1]{\text{$I(#1)$}}

\newcommand{\ranksum}[1]{$\text{\texttt{ranksum}}={#1}$}
\newcommand{\ignore}[1]{}

\newcommand{\myEqn}[1]{Eq. \ref{#1}}
\newcommand{\mySection}[1]{Section \ref{#1}}
\newcommand{\mySectionRange}[2]{Sections \ref{#1}--\ref{#2}}
\newcommand{\myFig}[1]{Fig. \ref{#1}}
\newcommand{\myTable}[1]{Table \ref{#1}}
 
\newcommand{\SMFig}[1]{supplementary Fig. SF{#1}}
\newcommand{\SMSection}[1]{supplementary Section S{#1}}
\newcommand{\SMSectionRange}[2]{supplementary Sections S{#1}--S{#2}}

\newcommand{\alphabet}{\mathbf{\aleph}}
\newcommand{\Mstate}{\texttt{m}}
\newcommand{\Istate}{\texttt{i}}
\newcommand{\Dstate}{\texttt{d}}
\newcommand{\Pmm}{\prob{\text{\Mstate} | \text{\Mstate}} }
\newcommand{\Pim}{\prob{\text{\Istate} | \text{\Mstate}} }
\newcommand{\Pdm}{\prob{\text{\Dstate} | \text{\Mstate}} }
\newcommand{\Pii}{\prob{\text{\Istate} | \text{\Istate}} }
\newcommand{\Pmi}{\prob{\text{\Mstate} | \text{\Istate}} }
\newcommand{\Pdi}{\prob{\text{\Dstate} | \text{\Istate}} }
\newcommand{\Pdd}{\prob{\text{\Dstate} | \text{\Dstate}} }
\newcommand{\Pmd}{\prob{\text{\Mstate} | \text{\Dstate}} }
\newcommand{\Pid}{\prob{\text{\Istate} | \text{\Dstate}} }

\usepackage{bold-extra}
\usepackage{xcolor}
\usepackage{hyperref}

\usepackage{graphicx}
\newsavebox\CBox
\def\textBF#1{\sbox\CBox{#1}\resizebox{\wd\CBox}{\ht\CBox}{\textbf{#1}}}

\begin{document}

\title{Bridging the Gaps in Statistical Models of Protein Alignment}

\author{%
\name{Dinithi Sumanaweera, Lloyd Allison$^*$ and Arun Konagurthu\thanks{Corresponding Authors.\\ Conceptualization: LA and ASK; Formal Analysis and Methodology: DS, LA and ASK; Software: DS;  Resources: ASK; Validation: LA and ASK; Visualization: DS; Writing: DS, LA and ASK; Supervision: LA and ASK; Project administration: ASK}}  % LA2: <- Is this of the "credit" system? 
%AK: Yes. I put these in. It could have been 'Lloyd's' credit system, since he independently suggested this to me long ago; alas he did not rush to publication and take credit for the credit system.
\address{Department of Data Science and Artificial Intelligence, Faculty of Information Technology, Monash University, Australia}
%\email{e-mail address}
}

\abstract{  % LA1: a rough cut ??...
This work demonstrates how a complete statistical model quantifying the evolution of pairs of aligned proteins can be constructed from a time-parameterised substitution matrix and a time-parameterised 3-state alignment machine. All parameters of such a model can be inferred from any benchmark data-set of aligned protein sequences. This allows us to examine nine well-known substitution matrices on six benchmarks curated using various structural alignment methods; any matrix that does not explicitly model a ``time''-dependent Markov process is converted to a corresponding base-matrix that does. In addition, a new optimal matrix is inferred for each of the six benchmarks.
Using Minimum Message Length (MML) inference, all 15 matrices are compared in terms of measuring the Shannon information content of each benchmark. This has resulted in a new and clear overall best performed time-dependent Markov matrix, \texttt{MMLSUM}, and its associated 3-state machine, whose properties we have analysed in this work. For standard use, the MMLSUM series of (log-odds) \textit{scoring} matrices derived from the above Markov matrix, are available at \url{https://lcb.infotech.monash.edu.au/mmlsum}.
\\~\\
%Intro
%\quickwordcount{Section:Introduction}
%Results
%\quickwordcount{Section:Results}
%Methods
%\quickwordcount{Section:Methods}

}

\date{\today}


\maketitle
\section{Introduction} \label{sec:intro}

Extant proteins diverge from their ancestors while tolerating considerable variation to their  amino acid sequences~\cite{chothialesk1986}. Inferring and comparing trustworthy relationships between their sequences is a challenging task and, when done properly, provides a powerful way to reason about the macromolecular consequences of evolution \cite{lesk2016book}. 

Many biological studies rely on identifying homologous relationships between proteins. The details of those relationships are represented as correspondences (alignment) between subsets of their amino acids. Each such correspondence suggests the divergence of the observed amino acids arising from a common locus within the ancestral genome. 

Although sequence comparison is a mature field, the fidelity of the relationships evinced by modern homology detection and alignment programs remains a function of the underlying models they employ to evaluate hypothesised relationships.  Most programs utilise a substitution matrix to quantify amino acid interchanges. These matrices are parameterised on a numeric value that accounts for the extent of divergence/similarity between protein sequences (e.g. PAM-250~\cite{dayhoff1978},  BLOSUM-62~\cite{henikoff1992}). 

Separately, the unaligned regions (gaps) of an alignment relationship are taken as insertions and deletions (\emph{indels}), accumulated during sequence evolution.
Separate gap-open and gap-extension penalties are widely felt to give plausible and sufficiently flexible models to quantify indels.
However, mathematically reconciling the quantification of substitutions with that of indels remains contentious.
Often the issue is simply avoided:
previous studies have shown that the choices of which substitution matrix to use, at what threshold of divergence/similarity, and with what values for gap penalties, remain anecdotal, sometimes empirical, if not fully arbitrary
\cite{vingron1994sequence,do2005probcons,sumanaweera2019statistical}. 
Evaluation of the performance of popularly-used substitution matrices is hampered by the fact that different sequence comparison programs yield conflicting results
\cite{do2005probcons,barton1987evaluation,vingron1994sequence,blake2001pairwise,loytynoja2008phylogeny}. Thus, as the field stands, it lacks an objective framework to assess how well the commonly-used substitution matrices perform for the task of comparison, without being impeded by ad hoc parameter choices. 
 
To address these lacunae, 
we describe an unsupervised probabilistic and information-theoretic framework that uniquely allows us to: 
\begin{enumerate}
    \item compare the performance of existing amino acid substitution matrices, in terms of Shannon information content, without any need for parameter-fiddling, 
    \item infer improved stochastic (Markov) models of amino acid substitutions that demonstrably outperform existing substitution matrices, and
    \item infer Dirichlet distributions accompanying the above Markov models, which provide a unified way to address amino acid insertions and deletions in probabilistic terms.
\end{enumerate}
    
Specifically, for any given collection of benchmark alignments, our framework uses the Bayesian Minimum Message Length (MML) criterion \cite{wallace2005statistical,allison2018coding} to estimate the Shannon information content \cite{shannon48} of that collection, measured in \emph{bits}. It is computed as the shortest encoding length required to compress losslessly all sequence pairs in the collection using their stated alignment relationships. This measure of Shannon information is based on rigorous probabilistic models that capture protein evolution (substitutions \textit{and} indels). Parameters are inferred unsupervised (i.e. automatically) by maximising the lossless compression that can be gained from the collection.  In general, Shannon information is a fundamental and measurable property of data, and has had effective use in studies involving biological macromolecules \cite{allison1990minimum,strait1996shannon,adami2004information}.

Central to our information-theoretic framework is a stochastic matrix describing a Markov chain, which models the probabilities of interchanges between amino acids as a function of time (of divergence). It works in concert with corresponding time-specific Dirichlet probability distributions that are learnt from the collection to model the parameters of the alignment finite-state machine, which is used to quantify the (information-measure of) complexity of alignments. This handling of alignment complexity over time-dependent 3-state ($\match, \insertion,$ and $\delete$) machine models overcomes any \emph{ad hoc} decisions about gap penalty functions and costs. Crucially, all probabilistic parameters in this framework are automatically inferred by optimising the underlying MML criterion. Using this framework we are also able to infer new, and demonstrably improved models of amino acid evolution. 

Furthermore, our framework enables the conversion of existing substitution matrices to corresponding stochastic matrices with high-fidelity, even those that do not explicitly model amino acid interchanges as a Markov process. This provides a way to directly and objectively compare the performance of substitution matrices without any parameter-tuning, and measure their respective lossless encoding lengths required to compress the same collection of benchmark alignments. To the best of our knowledge, all the above features are unique to the work presented here.

In the section below, we compare and contrast an extensive range of widely used substitution matrices introduced over the past four decades.
We demonstrate the improvement of a new substitution matrix (MMLSUM) inferred by our MML framework and analyse the properties of MMLSUM.  \mySection{sec:methods} gives a mathematical overview of our framework, with details explained in \SMSection{1}.

\section{Results and Discussion} \label{sec:results}

We consider nine well-known substitution matrices and new matrices inferred in this work.
In order of publication, the former are:
PAM~\cite{dayhoff1978}, JTT~\cite{jones1992rapid}, BLOSUM~\cite{henikoff1992}, JO~\cite{johnson1993structural}, WAG~\cite{whelan2001general}, VTML~\cite{muller2002estimating}, LG~\cite{le2008improved}, MIQS~\cite{yamada2013revisiting}, and PFASUM~\cite{keul2017pfasum}.
The matrices are compared on six different alignment benchmarks (\mySection{sec:methodsBenchmarks}) which are curated using diverse structural alignment programs.
Further, for each benchmark, our MML framework is able to infer a stochastic matrix that best explains the benchmark.
We compare the MML-inferred matrices against the existing matrices across the 6 benchmark collections.

The organization of this section is as follows:
\mySection{sec:resultsBenchmarks} describes the composition of the benchmarks.
\mySection{sec:results_msglens} presents the performance of all the matrices across each benchmark using the objective measure of Shannon information content when losslessly compressing the same benchmark alignment collection. From this we identify the most-generalisable matrix among the set of MML-inferred substitution matrices on various benchmarks, that outperforms the other matrices -- we term this matrix MMLSUM.  
\mySection{sec:analysisMMLSUM} analyses in detail MMLSUM, the best substitution model inferred here, for its characteristics including
aspects of physicochemical and functional properties that the matrix captures,
its relationship with expected change in amino acids as a function of divergence time, and
the properties of gaps that can be derived from the companion probabilistic models that work with MMLSUM,
amongst others.

\subsection{Composition of alignment benchmarks} 
\label{sec:2.1} \label{sec:resultsBenchmarks}

\begin{table}[!h] 
\caption{Composition of alignment benchmarks used in this work.}
\label{tab:benchmarks}
\scriptsize 
\begin{tabular}{ |c|c||r|r|r|r|r| }
 \hline
 \multicolumn{2}{|c||}{\textbf{Benchmark}} & 
  \multicolumn{1}{c|}{ \multirow{3}{*}{ \parbox{1.2cm}{ \textbf{Num. of seq.pairs}}}} 
 &
   \multicolumn{1}{c|}{ \multirow{3}{*}{ \parbox{1.2cm}{\centering \textbf{Num. of \\ \texttt{matches} }}}} 
 & 
    \multicolumn{1}{c|}{ \multirow{3}{*}{ \parbox{1.2cm}{\centering \textbf{Num. of \\ \texttt{inserts} }}}} 
 & 
     \multicolumn{1}{c|}{ \multirow{3}{*}{ \parbox{1.2cm}{ \centering \textbf{Num. of \\ \texttt{deletes}} }}} 
 & 
     \multicolumn{1}{c|}{ \multirow{3}{*}{ \parbox{1.2cm}{\centering \textbf{Avg. seq. \\ID} }}} 
 
 \\\cline{1-2} 
    \multicolumn{1}{|c|}{\multirow{2}{*}{ \parbox{2cm}{\centering \textbf{Name (Abbrv.)}}} }& 
 
    \multicolumn{1}{c||}{\multirow{2}{*}{ \parbox{3cm}{\centering \textbf{Curated with \\ structural aligner}}}}
 &  & & & &  \\ 
 &  & & & & & \\ 
 \cline{1-2} \cline{3-7} 
\hline \hline 
    \texttt{HOMSTRAD}     &  {MNYFIT, STAMP, COMPARER}   & 8323 &  1,311,478 & 96,911 & 98,810 &  35.1\%\\
    \texttt{MATTBENCH}    &  MATT                        & 5286  & 826,506 & 177,401 & 177,789 &  19.4\%\\
    \texttt{SABMARK-Sup}  &  SOFI, CE                    & 19,092 & 1,750,440 & 848,859 & 861,344 & 15.2\%\\
    \texttt{SABMARK-Twi}  &  SOFI, CE                    & 10,667 & 694,954 & 515,318 & 527,188 & 8.4\%\\
    \texttt{SCOP1}        &  DALI                        & 56,292 & 8,663,652 & 1,407,988 & 1,373,882 &  25.5\%\\
    \texttt{SCOP2}        & MMLIGNER                     & 59,092 & {8,145,678} & {1,673,687} & {1,653,531} & 24.8\%\\
        \hline
\end{tabular}
\end{table}

\myTable{tab:benchmarks} summarises the six alignment benchmarks in terms of the total number of sequence-pairs (and their corresponding alignments) that each collection contains, the number of observed $\{\match$, $\insertion$, $\delete\}$ states in their alignments, and their observed average sequence identity.\footnote{Sequence identify percentage of a pair of proteins is computed as the number of matched, identical amino acid pairs between two sequences divided by the length of the shorter sequence.}

% LA1:  Should there be citations (or hyperlinks) for the outside benchmarks (there are some in the .bib)???...
The distributions of sequence identity observed in each benchmark are shown in
\SMFig{2}. 
% LA1 pruning:   The representation of alignments in 
\texttt{HOMSTRAD} covers a wide range of sequence relationships. 
In comparison, 
\texttt{SABMARK-Sup} and \texttt{MATTBENCH} contain alignments of distant sequence-pairs. 
\texttt{SABMARK-Twi} contains alignments of sequences that have diverged into the `midnight zone' where the detectable sequence signal is extremely feeble.
Finally, the largest benchmark we use contains 59,092 unique sequence-pairs sampled from the superfamily and family levels of SCOP~\cite{murzin1995scop}. These pairs were aligned separately using DALI\footnote{Of the 59,092 pairs in the sampled SCOP data-set, DALI does not report any alignment for 2800 pairs.} \cite{DALI} and MMLigner \cite{collier2017statistical} structural alignment programs to obtain \texttt{SCOP1} and \texttt{SCOP2} benchmarks, respectively (\mySection{sec:methodsBenchmarks}).

Collectively, all six benchmarks cover varying distributions of sequence relationships,  whose alignments were curated using diverse structural alignment programs (second column of \myTable{tab:benchmarks}). This diversity of chosen benchmarks minimises the possibility of introducing any systematic bias to the evaluation of models of amino acid substitution.

\subsection{Shannon information content of benchmarks}
\label{sec:results_msglens} \label{sec:resultsShannonInfo}

The lossless encoding length is estimated for each benchmark using the MML framework described in \mySection{sec:methods}. The framework quantifies the Shannon information content of the benchmark, measured in \emph{bits}, under varying models of amino acid substitution.
That is, for each benchmark, the encoding scheme has a choice of either employing an existing substitution matrix (in its stochastic matrix form -- see section \mySection{sec:matrixConversion}), or automatically inferring a new stochastic matrix optimal to that collection. 

Further, using the notations described in \mySection{sec:mmlframework}, for a stochastic matrix $\submat{}$ chosen to losslessly compress all the sequence-pairs in a specific alignment benchmark  $\data{}$, all the other models involved in this MML information-theoretic framework, i.e. $\{\nullmodel, \dirparams, \stateParams, \tau\}$ (\mySection{sec:mmlframework}), are automatically-inferred (optimised) for  $\submat{}$ on each benchmark $\data{}$ under the MML criterion.

\myTable{tab:totalResults1} presents the lengths of the shortest encoding (i.e. Shannon information) using each of the 15 matrices (9 existing; 6 inferred) to explain each of the six benchmarks.
\newcommand{\mycol}[1]{\textcolor{blue}{\textBF{#1}}}
\begin{table}[t]
    \caption{Shannon information content (in bits) to losslessly encode each benchmark by varying the substitution matrices. The rank order of their performance is reported in each column within parentheses (smaller lossless encoding length = better rank). Best encoding length for each benchmark is highlighted in bold blue font. All other parameters are inferred (optimised) by the MML framework for the matrix chosen to losslessly compress each benchmark.}
    \label{tab:totalResults1}
\scriptsize
    \centering
    \begin{tabular}{|@{~}l@{~}||@{~}r@{~}|@{~}r@{~}|@{~}r@{~}|@{~}r|@{~}r@{~}|@{~}r@{~}||}
    \hline 
\multicolumn{1}{|@{~}c@{~}||}{\textbf{Benchmark \shortstack{~\\$\rightarrow$\\ (D)}}} &  
\multicolumn{1}{|c|}{\texttt{\textbf{HOMSTRAD}}} &  
\multicolumn{1}{|c|}{\texttt{\textbf{MATTBENCH}}} &  
\multicolumn{1}{|c|}{\texttt{\textbf{SABMARK-Sup}}} & 
\multicolumn{1}{|c|}{\texttt{\textbf{SABMARK-Twi}}}  &  
\multicolumn{1}{|c|}{\texttt{\textbf{SCOP1}}}  & 
\multicolumn{1}{|c|}{\texttt{\textbf{SCOP2}}} \\ 
    \hline \hline 
    \multicolumn{1}{|@{~}c@{~}||}{Matrix ($M$) $\downarrow$}&
    \multicolumn{6}{|c|}{Shannon information content using existing substitution matrices (and its rank across all matrices)} \\
    \hline \hline
\texttt{PAM} (1978)                            &   11531556.4 (15) & 9143136.9 (15)  &   23574085.5 (15) &  11310226.0 (15)  & 84925406.9 (15)   &     82757945.5 (15) \\
\texttt{JTT} (1992)                            &   11481203.2 (14) & 9072068.6 (13)  &   23450831.6 (13) &  11251914.3 (13)  & 84353986.5 (13)   &     82218532.0 (13)\\
\texttt{BLOSUM} (1992)                         &   11437552.8 (10) & 9037049.8 (08)  &   23373908.1 (07) &  11228043.6 (06)  & 84174710.8 (11)   &     81995179.3 (10)\\ 
\texttt{JO} (1993)                             &   11476518.5 (13) & 9118266.0 (14)  &   23501361.5 (14) &  11290056.3 (14)  & 84567477.8 (14)   &     82405562.0 (14) \\
\texttt{WAG} (2001)                            &   11419186.0 (05) & 9052722.9 (12)  &   23400017.0 (11) &  11243242.0 (12)  & 84141633.8 (09)   &     81996154.3 (11)\\
\texttt{VTML} (2002)                           &   11423498.2 (07) & 9035903.4 (07)  &   23377505.0 (08) &  11230624.1 (08)  & 84075908.5 (07)   &     81925302.7 (07)\\
\texttt{LG}  (2008)                            &   11464263.6 (12) & 9049040.6 (11)  &   23411713.3 (12) &  11235389.2 (09)  & 84255656.9 (08)   &     82090289.0 (12)\\
\texttt{MIQS} (2013)                           &   11422215.4 (06) & 9040480.8 (10)  &   23385242.8 (10) &  11236323.3 (10)  & 84076742.8 (12)   &     81927707.6 (08) \\
\texttt{PFASUM} (2017)                         &   11412888.2 (02) & 9039799.4 (09)  &   23379074.4 (09) &  11236572.3 (11)  & 84040519.3 (04)   &     81902713.6 (06)\\
    \hline \hline 
    \multicolumn{1}{|@{~}c@{~}||}{Matrix ($M$) $\downarrow$}&
    \multicolumn{6}{|c|}{Shannon information using MML-inferred matrices (and its rank across all matrices)}\\
    \hline \hline
    
MML$_\text{\texttt{HOMSTRAD}}$       & \mycol{11405604.7 (01)}   &  9035317.6 (05)  &  23365151.2 (06)  & 11230184.9 (07)   &  84026302.3 (03)  &  81873575.9 (03)    \\
MML$_\text{\texttt{MATTBENCH}}$      & 11426344.1 (09) &  \mycol{9025882.4 (01)}  &  23355215.8 (03)  & 11217219.1 (03)  &  84050927.4 (05)  &  81886796.3 (04)     \\
MML$_\text{\texttt{SABMARK-Sup}}$    & 11424135.9 (08) &  9031315.9 (04)&  \mycol{23346025.4 (01)}  & 11212252.7 (02)   &  84067892.9  (06) &  81889152.3 (05)    \\
MML$_\text{\texttt{SABMARK-Twi}}$    & 11442781.7 (11) &  9035720.5 (06)&  23356054.5 (05) & \mycol{11211360.9 (01)}   &  84155701.5 (10)  &  81962307.9  (09)   \\
MML$_\text{\texttt{SCOP1}}$         & 11413295.6 (03) &  9029682.8 (03) &  23355235.9 (04) & 11221295.6  (05) &  \mycol{83996796.0 (01)}  &  81848381.0 (02)    \\
MML$_\text{\texttt{SCOP2}}$         & 11413725.3 (04) &  9028667.52 (02)  &  23349826.8 (02) & 11218205.6 (04)  &  83999536.4 (02)  &  \mycol{81840654.6 (01)}     \\
\hline
    \end{tabular}

\end{table}
Previously published  matrices are arranged in chronological order of publication (rows). The last five rows show results for the stochastic matrices  inferred from each benchmark.

To get an overall view of the performance of each matrix as a consensus over all benchmarks generated from individual ranks (shown within parentheses in each column of \myTable{tab:totalResults1}), we use a simple-yet-effective statistic: the (row-wise) sum of ranks of each matrix over all benchmarks, \texttt{ranksum} in short. Since this evaluation involves ranking 15 matrices over 6 benchmarks, the \texttt{ranksum} of any matrix is an integer between $6\times 1 = 6$ (best possible performance) and $6\times 15 = 90$ (worst possible performance).  

Among the set of existing matrices, \texttt{PAM} (\ranksum{90}) consistently gave the worst (i.e. longest) lossless encoding lengths across all benchmarks. This is anticipated, as \texttt{PAM} was derived in 1978 using the then available set of alignments. This is followed by the performance  of \texttt{JO} (\ranksum{83}), \texttt{JTT} (\ranksum{79}), \texttt{LG} (\ranksum{64}), \texttt{WAG} (\ranksum{60}), \texttt{MIQS} (\ranksum{56}), \texttt{BLOSUM} (\ranksum{52}), \texttt{VTML} (\ranksum{44}) and \texttt{PFASUM} (\ranksum{41}). From these numbers it can be seen that, by and large, the previously published models of amino acid substitutions have improved over time. \texttt{BLOSUM} is among the earliest matrices (published in 1992) that outperforms several matrices that were proposed much later, and is only superseded in performance by \texttt{VTML} (published in 2002) and \texttt{PFASUM} (published recently in 2017), among the later matrices.

In comparison, the (stochastic) matrices inferred by our framework, specific to each benchmark, perform consistently better than previously-published substitution matrices. Indeed it is to be expected that the encoding length of any MML-inferred matrix that was optimised on a specific benchmark will outperform all other matrices on that benchmark -- and this is precisely what is observed in \myTable{tab:totalResults1} (see highlighted terms). However, the utility of any matrix lies in its ability to generalise to \textit{other} benchmarks and perform well on those. The table above clearly demonstrates the ability of MML-inferred matrices to generalise and explain other benchmarks, far outperforming all existing ones. From the point of view of their \texttt{ranksums}: MML$_\texttt{SABMARK-Sup}$ (i.e., the stochastic matrix inferred on SABMARK-Sup benchmark) gives \ranksum{26} across all benchmarks, while  MML$_\texttt{MATTBENCH}$ and MML$_\texttt{HOMSTRAD}$ give \ranksum{25}. The top two performers overall come from the matrices inferred on the two  SCOP benchmarks,  MML$_\texttt{SCOP1}$ (\ranksum{18}) and MML$_\text{SCOP2}$  (\ranksum{15}). 

The sole outlier among the MML-inferred matrices was MML$_\text{SABMARK-Twi}$ with \ranksum{42}. As already stated (\mySection{sec:resultsBenchmarks}),  \texttt{SABMARK-Twi} benchmark contains alignments of highly-diverged sequence-pairs (avg. seq. identity of 8.4\%). Thus, the benchmark itself provides an extremely weak sequence signal to infer a stochastic matrix that can be generalised effectively to explain a wider range of sequence relationships that other benchmarks embody. But a noteworthy observation is that MML$_\texttt{SABMARK-Twi}$ (\ranksum{42}) is nearly in par with \texttt{PFASUM} (\ranksum{41}) which was the best performer among the set of existing matrices. See \SMSectionRange{2}{3} for an extended analysis and additional information. 

 Overall, the MML-inferred matrix from the \texttt{SCOP2} benchmark (MML$_{\text{\texttt{SCOP2}}}$) with a \ranksum{15} outperforms all other matrices.
 The is because \texttt{SCOP2} benchmark is three times larger than \texttt{SABMARK-Sup} (seven times that of \texttt{HOMSTRAD}) and contains a wider range of sequence relationships than other benchmarks. Thus, all of our subsequent analyses will involve MML$_{\text{\texttt{SCOP2}}}$ matrix -- we will refer to it as MMLSUM (for MML subtitution matrix).  % LA1: I always wondered why the name!-)

\subsection{Analysis of MML substitution matrix (MMLSUM)}
\label{sec:analysisMMLSUM}

\begin{figure}[!h]
    \centering
    \includegraphics[width=\textwidth]{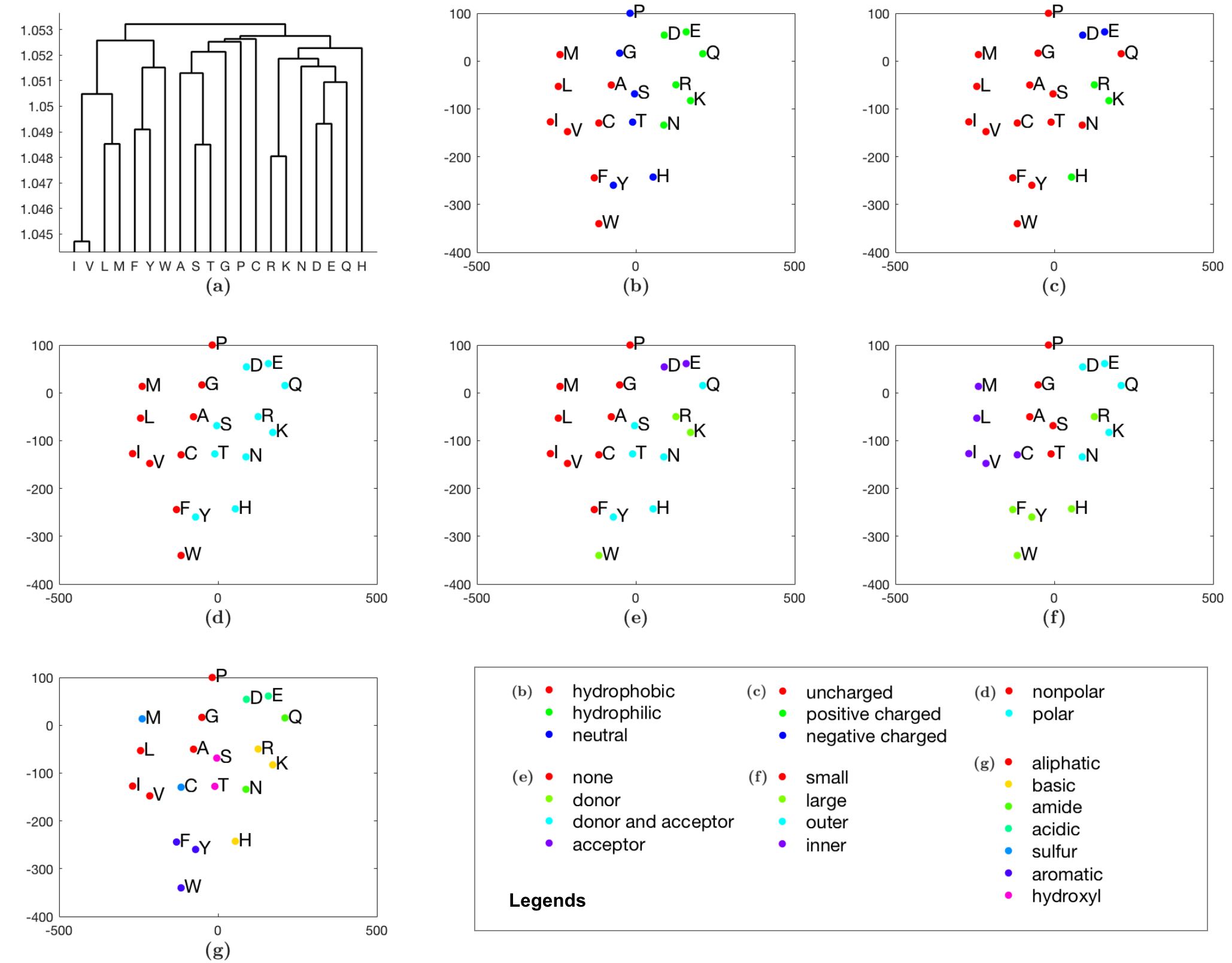}
    \caption{(a) Average-linkage clustering of amino acids generated from MMLSUM. (b)-(g) tSNE clustering of amino acids generated from MMLSUM. All plots have the same clustering, but coloured under different amino acid classification schemes based on: (b) hydropathy,  (c) charge, (d) polarity, (e) Hydrogen donor or acceptor, (f) Swanson classification of amino acids~\cite{swanson1984unifying} and (g) chemical  properties (based on IMGT classification \cite{lefranc2015imgt}). Refer to the legend for the colouring scheme in various subplots.
    \ignore{-- \url{http://www.imgt.org/IMGTeducation/Aide-memoire/_UK/aminoacids/IMGTclasses.html})}}
    
    \label{fig:clusters}
\end{figure}

\subsubsection*{Amino acid clustering}
 
Here we analyse groupings of amino acids implicit in MMLSUM.  \myFig{fig:clusters}(a) gives a dendrogram generated using the average-linkage method on the MMLSUM base stochastic matrix (at $t=1$). 
We notice several important clusters that have been previously flagged as necessary for any reliable matrix quantifying amino acid substitutions~\cite{dayhoff1978}.  The following groups can be identified:
\begin{itemize}
    \item Hydrophobic amino acids Valine (V), Isoleucine (I), Leucine (L),  and Methionine (M) cluster into a distinct clade.
    \item Aromatic amino acids Tryptophan (W), Tyrosine (Y), and Phenylalanine (F) group into another clade. 
    \item Neutral amino acids Alanine (A), Serine (S), Threonine (T), Glycine (G) and Proline (P) group together. 
    \item Large amino acids Arginine (R), Lysine (K), Asparagine (N), Aspartic acid (D), Glutamic acid (E), and Glutamine (Q) form a clade.
    \item The remaining two amino acids Histadine (H) and Cysteine (C) cluster apart from the rest. 
\end{itemize}

To study the groupings observed more systematically and from a different point of view, we  apply the technique of t-distributed stochastic neighbor embedding (tSNE)~\cite{maaten2008visualizing} to MMLSUM.
tSNE performs a non-linear dimensionality reduction of high-dimensional feature space and gives visualizations that aid detection of clustering in lower dimensions~\cite{platzer2013visualization}.  \myFig{fig:clusters}(b)-(g) all show the same two-dimensional tSNE-visualisation of amino acids from MMLSUM, but each subplot colours the amino acids differently, based on widely-used amino acid classification schemes. 
These different schemes encompass the hydropathic character of amino acids, their charge, their polarity, their donor/acceptor roles in forming hydrogen bonds, their size and propensity for being buried/exposed, and their chemical constitution.

In \myFig{fig:clusters}(b)-(f), the visualization yields clearly separable amino acid groups on tSNE's 2D embedding of MMLSUM. In \myFig{fig:clusters}(g) which deals with the classification based on the chemical characteristics of amino acids (as per IMGT~\cite{lefranc2015imgt}) the classes are mostly well-differentiated, barring a few outliers that include  Histadine (H), Cysteine (C) and Asparagine (N) -- we note that  H and C were also outliers in the hierarchical clustering (cf. \myFig{fig:clusters}(a)).

\subsubsection*{Expected amino acid change and properties of gap lengths as a function of time of divergence}

\begin{figure}[!b]
    \centering
    \includegraphics[width=\textwidth]{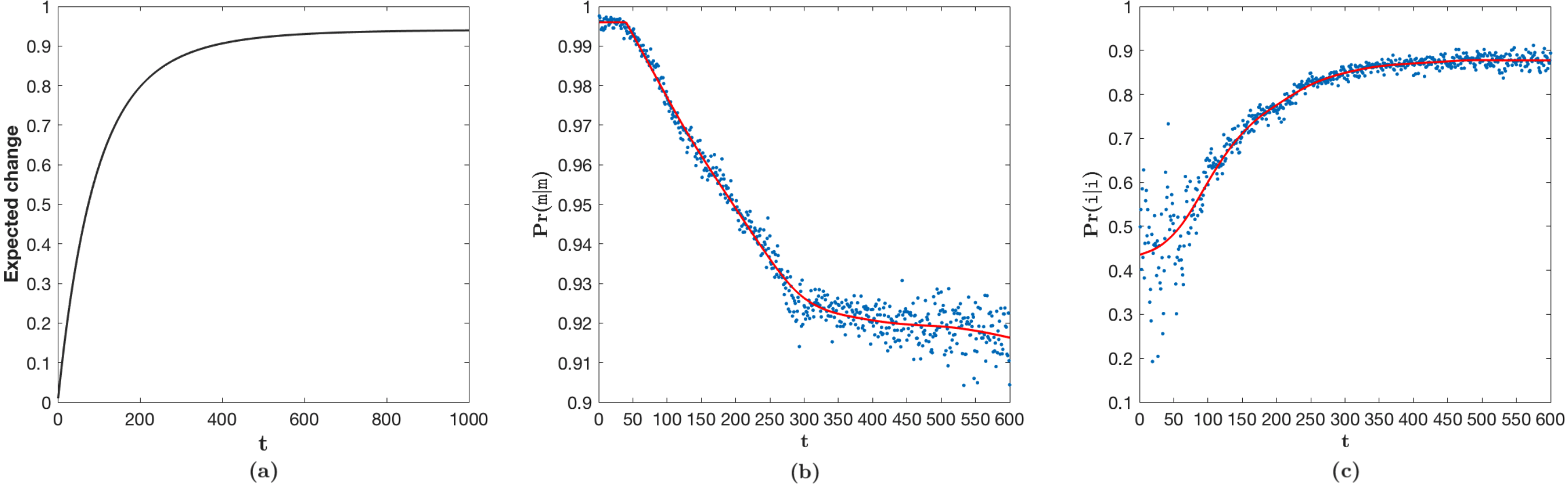}
    
   \caption{(a) Expected change of amino acids under the MMLSUM's model as a function of divergence-time parameter $t$ (b) The variation of $\Pmm$ when derived from the \textit{mean} of the inferred time-dependent Dirichlet distributions accompanying MMLSUM. (c) Similarly, the variation of $\Pii$ estimate with $t$. The divergence time parameter $t$ is plotted in the range $[1,600]$, beyond which the amino acids have near-converged to the stationary distribution of MMLSUM (See \SMFig{6})}
    \label{fig:mmlsumKL}
\end{figure}

\myFig{fig:mmlsumKL}(a) shows the growth of expected change of amino acids implicit in MMLSUM as a function of time $t$.
Previous studies~\cite{rost1999twilight} have shown that protein sequence relationships are most reliable when their sequence identity is $>40\%$ (or expected amino acid change $<60\%$). This corresponds approximately to the range $t\in[1,100]$ in \myFig{fig:mmlsumKL}(a). The `twilight zone' of sequence relationships has been characterised by relationships sharing $[20-35]\%$ sequence identity (or $[65-80]\%$ change). This corresponds approximately to the range $t\in[125-200]$. Expected change of $90\%$ is reached at $ t = 400 $ and increases very slowly thereafter ($94\%$ change at $t=1000$). 
(An extended analysis from the point of view of Kullback-Leibler divergence of individual amino acids with respect to the stationary distribution of MMLSUM is presented in \SMSection{3.3}.)

Further, unique to this framework is the optimal inference of ``time''  ($t$) dependent Dirichlet distributions  that work in concert with the inferred stochastic matrix, MMLSUM. These distributions model time-specific state transition probabilities of the alignment 3-state machine over \texttt{match (m)}, \texttt{insert (i)}, and \texttt{delete (d)} states (see \myFig{fig:fsm1}).  \mySection{sec:dirichletIntro} introduces the 9 transition probabilities involved in the alignment 3-state machine, of which three are free  ($\Pmm$, $\Pmi$, and $\Pii$), and the remaining dependent.

In the 3-state machine, the probability of moving from a \texttt{match} state to another \texttt{match} state ($\Pmm$) controls the run length of any block of matches in an alignment. The expected value of this run length, a geometrically distributed variable, is given by $\frac{1}{1-\Pmm}$. Also, the value $1-\Pmm$ gives the probability of a gap (i.e, a block of insertions or deletions of any length) starting at a given position  
in an alignment.

\myFig{fig:mmlsumKL}(b) plots the values of $\Pmm$
derived from the mean values 
of the inferred Dirichlets for the \match\ state.
We observe that it remains nearly a constant ($\Pmm \approx  0.9958$) in the range of $t\in[1,40]$. This value corresponds to an \emph{expected} run length of  $\sim238$ amino acids per block of matches. Sequence-pairs whose time parameter is in that range are closely-related, with $>67\%$ amino acids expected to be \emph{conserved} (cf. \myFig{fig:mmlsumKL}(a)).
The probability  of opening a gap
($1-\Pmm=0.0042$)  
for sequence-pairs in this range is extremely small. Next, in the range $t\in[40,300]$ $\Pmm$ decreases linearly with $t$. Comparing this range in \myFig{fig:mmlsumKL}(a), the expected change of amino acids drastically increases from $\sim32\%$ to $\sim87\%$. This correlates with the expected length of match-blocks dropping from $233$ amino acid residues to about $13$.  Further, for $t\ge 300$, $\Pmm$ decreases only gradually. 

Similarly, the free parameter $\Pii$ (equalling $\Pdd$ in the \emph{symmetric} alignment state machine)  controls the run lengths of indels. \myFig{fig:mmlsumKL}(c) gives values of $\Pii$ derived from the mean values of the inferred Dirichlets for the \insertion\ state. In the range $t\in[1,50]$ values of $\Pii$ are noisy because the probability of observing a gap is small. Hence, there are only few observations of gaps from which to estimate this parameter.  
However, in the range of $t\in[50,400]$ $\Pii$ grows from $0.5248$ to about $0.8431$, beyond which the probability flattens out at about $0.8759$ on average (expected gap length $\frac{1}{1-\Pii}=\sim 8$ amino acid residues). 
The change of $\Pmi$ with $t$ mirrors the behaviour of $\Pii$. 
This is because $\Pii+\Pmi+\Pdi=1$, and $\Pdi$ remains very small.

\subsubsection*{Function similarity and evolutionary distance}
\label{sec:2.5} \label{sec:resultsFunction}

We analyse how functional similarity between the protein domain-pairs in the \texttt{SCOP2} benchmark correlates with the automatically-estimated time (of divergence) parameter, under the MMLSUM model, for each of its $59,092$ sequence-pairs. (Refer \SMFig{5} for the distribution of the inferred set of time parameters.)
For this, we employ the Gene Ontology (GO)~\citep{gene2004gene} that provides function annotations for protein domains in three categories: (1) the `Biological Process' (BP) they come from, (2) the `Molecular Function' (MF) they exhibit, and (3) the `Cellular Component' (CC) they belong to. Due to missing tags in the GO database, not all pairs could be considered for the analysis presented below:  % LA2:  a cut 
we considered only those domain-pairs where both their domains have one or more of the above categories tagged in the GO database. This resulted in 37,201 pairs for exploration of functional similarity at the level of BP, 48,215 pairs at the level of MF, and 31,594 at the level of CC.

\begin{figure}[!t]
\centering 
\includegraphics[scale=0.3]{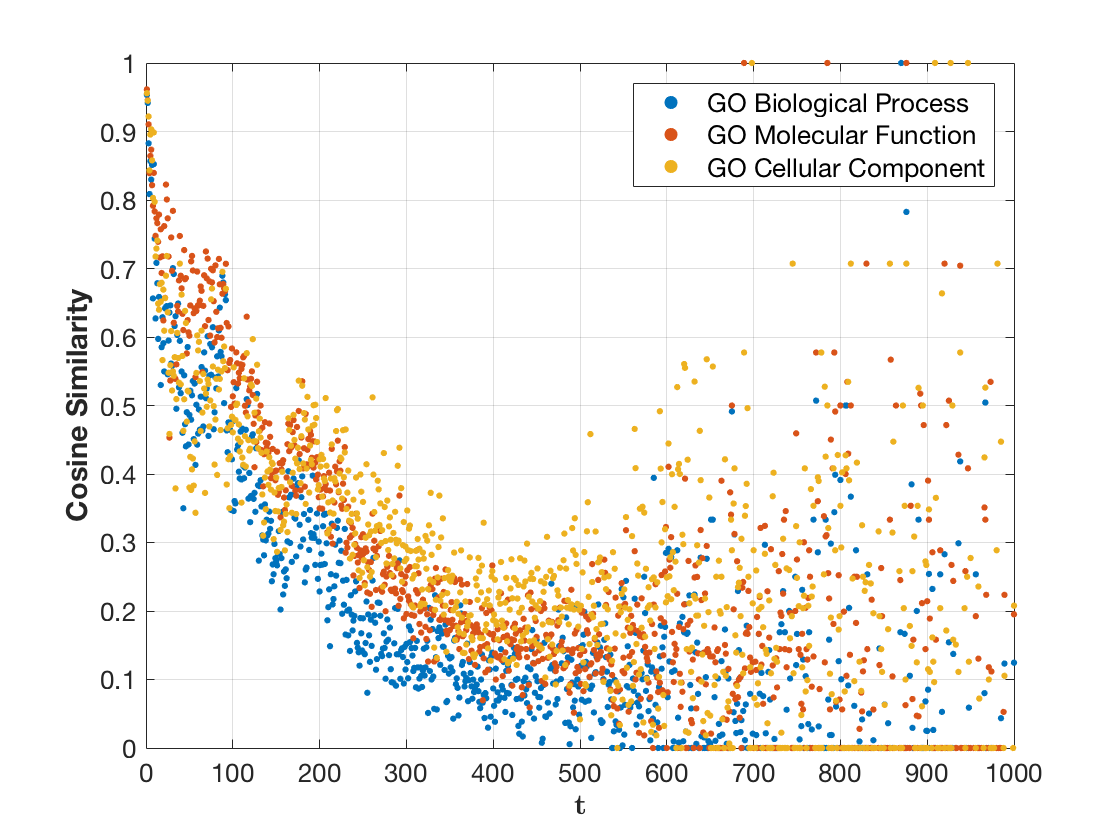}
\caption{How the similarity of the Gene Ontology annotation vectors between  two sequences of a structurally aligned pair in the \texttt{SCOP2} benchmark varies as a function of the optimal evolutionary distance $n$, under the MMLSUM substitution model.}
\label{fig:functionSim}
\end{figure}

The function-similarity between a domain pair is evaluated on a similarity measure involving the list of terms within each category as follows. Each domain is represented as a boolean vector corresponding to the observed set of distinct terms in the GO database. For each domain-pair, two such vectors $\vec{x}$ and $\vec{y}$) are constructed and their cosine similarity, $\frac{\vec{x}\cdot\vec{y}}{||\vec{x}|| ||\vec{y}||}$, is computed. \myFig{fig:functionSim} plots the average changes of this measure as a function of $t$.

Overall, the observed trend seen in \myFig{fig:functionSim} correlates with the plot showing the convergence of amino acids to stationary distribution of MMLSUM
(cf. \SMFig{6})
As expected, the function-similarity measure decreases as the domains diverge from each other, and thereby pick up new functions. The similarity measure flattens out
(and becomes noisy) for $t>400$.  

Interestingly, studying the `phylum' (taxonomic rank) of each domain reveals the divergence of function from another point of view.  We analysed the proportion of \texttt{SCOP2} domain-pairs that both belong to the same phylum by binning their inferred time parameters using MMLSUM. We find that 92.3\% of the domain-pairs whose inferred time parameters are in the range $t\in[1,50]$ belong to the same phylum. Between $t\in(50,100]$ this proportion falls to 53.5\%. We observe roughly similar proportion, 50.4\%, for values of $t\in(100,200]$. Between $t\in(200,300]$ and $t\in(300,600]$ the number drops more drastically to 34.1\% and 32\%, respectively.

\subsection{Conclusion} \label{Conclusion}
\label{sec:resultsConclusion}

A time-dependent substitution matrix and an associated 3-state alignment machine were combined to give a unified statistical model of aligned protein sequences (\mySection{sec:methods} and \SMSection{1}). This model uniquely provides many advantages including the inference of evolutionary \emph{distance} 
between two aligned protein sequences (\SMSection{1.3.3}), the estimation of Shannon information content of any benchmark data-set containing a collection of alignments (\mySection{sec:mmlframework}), and using this to compare between previously published substitution matrices (\mySection{sec:results_msglens}).

Existing substitution matrices that do not explicitly depend on time were each converted to a corresponding base matrix that \textit{can} benefit from our information-theoretic framework.
Optimal matrices were also inferred using MML for each of six benchmark data-sets of aligned proteins. This allowed us  to compare nine existing substitution matrices with the six MML-inferred matrices, on six benchmarks.
All MML-inferred matrices perform very well and, in particular, the MMLSUM matrix outperforms all of the other matrices and generalises best of all, across all six benchmarks.
The MMLSUM matrix implies sensible groupings of the 20 amino acids (\mySection{sec:analysisMMLSUM}).
Increasing evolutionary distance correlates with decreasing similarity of function in aligned proteins, becoming noisy after $t=400$.   
The complete statistical model yields an interesting relationship 
between evolutionary distance and the frequency and length of gaps (indels).
We have made available the MMLSUM series of log-odds \emph{scoring} matrices for standard use at \url{https://lcb.infotech.monash.edu.au/mmlsum}.

\section{Materials and methods}
\label{sec:methods} \label{sec:methods}

\subsection{Introduction to Minimum Message Length framework}
\label{sec:mmlintro} \label{sec:introMML}

The Minimum Message Length (MML) principle \cite{wallace1968information, wallace2005statistical, allison2018coding} is a powerful technique to infer reliable hypotheses (models, theories) from observed data. MML is an information-theoretic criterion that, in its mechanics, combines Bayesian inference~\cite{bayes1763lii}
with lossless data compression.  
Formally, the joint probability of any hypothesis $H$ and data $D$ is given by
$$
    \prob{H,D} = \prob{H}\prob{D|H} = \prob{D}\prob{H|D}
$$
Commonly, model inference depends on identifying suitable hypotheses based on posterior probability (i.e.,  $\prob{H|D}$ --  the probability of any hypothesis given the data).  Separately, Shannon’s Mathematical Theory of Communication \cite{shannon48} quantifies the amount of information in any event $E$ that occurs with a probability of $\prob{E}$ as:
\begin{align*}
   \I{E} = -\log_2{\prob{E}} \quad \text{bits}
\end{align*}
$\I{E}$ can be understood as the minimum lossless encoding length required to communicate the event E. Using this, the joint probability $\prob{H,D}$ can in turn be expressed in terms of Shannon information content as:
\begin{equation} \label{eq:1}
    \I{H,D} = \I{H} + \I{D|H} = \I{D} + \I{H|D}
\end{equation}
This relationship can be rationalised as the length of a two-part message required to communicate the hypothesis $H$ and the data $D$ given $H$, ($D|H$), as a notional communication between a transmitter and receiver. In this formulation, the transmitter losslessly encodes an hypothesis $H$ which takes $\I{H}$ bits to state, followed by the data D given the stated hypothesis H, taking another $\I{D|H}$ bits to state.  Note, for any $H$ and $D$, $\I{H}$ and $\I{D|H}$ have to be accurately estimated, and we carry out this estimation using the well-established technique from the statistical-learning literature, due to  \citet{wallace1987estimation}.  
Note, one of the important aspects of MML is stating (i.e., lossless encoding) parameter estimates to optimum precision.

Many attractive properties emerge from the MML formulation, but most useful here is the observation that the difference in message lengths between any pair of competing hypotheses (say $H_1$ and $H_2$) gives the posterior log-odds ratio:
\begin{align*}
 \I{H_1,D} - \I{H_2,D} &= 
   -\log_2{\prob{D}\prob{H_1|D}} + \log_2{\prob{D} \prob{H_2|D}} \\
                &= \log_2{\left( \frac{\prob{H_2|D} }{\prob{H_1|D} }  \right)  }  
\end{align*}
This allows competing hypotheses to be objectively compared and the best one reliably chosen.

\subsection{Formulating the problem in the MML framework}
\label{sec:mmlframework}
In this work, the observed data $\data$ denotes any data-set of aligned protein sequences. Formally, it is composed of pairs of amino acid sequences and their \textit{given} alignments: 
\begin{align*}
    \data = \set{ \alseqpair{1}, \alseqpair{2}, ... , \alseqpair{|\mathbf{D}|} }
\end{align*}
where each $\seqS{i}$ and each $\seqT{i}$ is a sequence over the alphabet of 20 amino acids, and $\alignment{i}$ represents their given alignment relationship specified as a 3-state string over $\{$\match, \insertion, \delete$\}$ states. Note that each alignment, $\alignment{i}$, is a part of the observed data, coming from structural alignments or from some set of benchmark alignments -- see \mySection{sec:methodsBenchmarks} for details.

A hypothesis $H$ that losslessly explains the above data $\data$ is composed of the following statistical models (and we emphasize all the models shown below are automatically inferred from any given $\data$, as those that are optimal under the MML criterion):

\begin{enumerate}
    \item A stochastic Markov matrix $\submat{}$ (see \mySection{sec:stochasticMatrixIntro}) which is used to losslessly encode the corresponding pairs of amino acids in $\seqpair{i}$ that are under ‘\match’ states in $\alignment{i}$.
    Note, this stochastic matrix can either be optimally inferred under the MML criterion from the collection $\data$ (see \mySection{sec:inferBestMatrix}), or, for comparison purposes, be any existing substitution matrix (see \mySection{sec:matrixConversion}).
    
    \item A multinomial model, $\nullmodel$, of the 20 amino acids, used to losslessly encode the amino acids in the unaligned regions of $\seqpair{i}$, i.e., those that are under insert or delete (indel) states in $\alignment{i}$.     Estimates of $\nullmodel$ are optimally inferred using MML from the indel regions observed in alignments in the collection $\data$. (Note, for a comparison, we also explore two other choices for $\nullmodel$: one derived from the stationary distribution of $\submat{}$, and the other derived from a source independent of $\data$ (refer \SMSection{2.2})). 
    
    \item A set of automatically inferred Dirichlet parameters $\dirparams$ 
    (see \mySection{sec:dirichletIntro}), each one specifying a Dirichlet distribution for a specific value of ``time'', $t$,
    to be used in conjunction with $\submat{}(t)$;
    the alignment 3-state machine’s transition probabilities $\stateParams_i$ that is inferred optimally for any alignment $\alignment{i}$ are encoded using one of the time-dependent Dirichlet distributions.
    These transition probabilities in turn are used to losslessly encode a 3-state alignment string $\alignment{i}$.
    
    \item Finally, the set of automatically inferred time parameters 
    $\boldsymbol{\tau}=\set{t_1,t_2, ... , t_N}$, one for each sequence-pair in D, using the above models. Each $t_i$ captures the divergence of corresponding sequence-pairs $\seqpair{i}$, where $t_i$ can be interpreted as the length of the Markov chain by which their amino acids are related using the above models. 
\end{enumerate}
Using \myEqn{eq:1} (see \mySection{sec:mmlintro}), this framework allows the estimation of the Shannon information content in the hypothesis $H$ and data $\data$ as a summation of individual Shannon information terms: 
    \begin{equation} \label{eq:2} \displaystyle 
        \I{H,\data} = \I{\submat{}} + \I{\nullmodel} + \I{\dirparams} + 
        \sum_{i=1}^{|\mathbf{D}|} \I{t_i} + \I{\stateParams_i | \dirparams,t_i} + 
        \I{\alignment{i}|\stateParams_i,t_i} + \I{\seqpair{i}|\alignment{i},\submat{},\nullmodel,t_i}
    \end{equation}
where, \\ 
$\I{\submat{}}$ is the lossless encoding (i.e., statement) length of Matrix $\submat{}$ that models matched parts of $\data$; \\
$\I{\nullmodel}$ is the  statement length of probability estimates $\nullmodel$ to model indel parts of $\data$; \\
$\I{\dirparams}$ is the statement length of inferred time-dependent Dirichlet parameters; \\
$\I{t_i}$ is the statement length of inferred time $t_i$ of a sequence-pair $\seqpair{i}$ given its alignment $\alignment{i}$; \\
$\I{\stateParams_i | \dirparams,t_i}$ is the statement length of alignment state machine parameters inferred on each $\alignment{i}$; \\
$\I{\alignment{i}|\stateParams_i,t_i} $ is the statement length of each $\alignment{i}$; and \\
$\I{\seqpair{i}|\alignment{i},\submat{},\nullmodel,t_i}$ is the statement length of explaining all amino acids in the sequence-pair. \\

The models stated above and their MML estimation are briefly described in \mySectionRange{sec:methodsBenchmarks}{sec:mmlTermsEstimation}, whereas the full details of their inference are presented in the \SMSection{1}.

\subsection{Alignment benchmarks and materials} 
\label{sec:methodsBenchmarks} 

This work utilises the following benchmarks to validate the framework introduced above. Each benchmark individually provides a source collection $\data$ of pairs of sequences and their given alignment relationships. $\data$ is losslessly compressed under the minimum message length (MML) criterion, \myEqn{eq:2}. 

\begin{enumerate} 
    \item \underline{HOMSTRAD} \cite{homstrad} (\url{https://mizuguchilab.org/homstrad}) is a database of structural alignments for homologous protein families. It contains multiple alignments of proteins covering 1032 families with known structures. Their alignments are semi-manually curated using the structural alignment programs: MNYFIT, STAMP and COMPARER \cite{sutcliffe1987knowledge,russell1992multiple,sali1990definition}. 
    
    \item \underline{Mattbench} \cite{daniels2011touring} (\url{https://bcb.cs.tufts.edu/mattbench/Mattbench.html}) curated using the structural alignment program MATT \cite{menke2008matt}. This work combines into one benchmark, its two sets of alignments classified as \emph{superfamily} and \emph{twilight zone}. The superfamily set contains alignments of 225 groups of homologous protein domains, where all pairs of domains in any group have a sequence identity $<50\%$.  The twilight zone set is a much smaller and distinct set containing alignments covering 34 distantly related groups, where the sequence identity threshold is $<20\%$ \cite{daniels2011touring}.

    \item \underline{SABMARK} \cite{van2005sabmark} (\url{http://bioinformatics.vub.ac.be/databases/databases.html}) is a more extensive set of alignments covering \emph{superfamily} and \emph{twilight zone} protein domain sets, whose alignments are curated using SOFI and CE \cite{boutonnet1995optimal,shindyalov1998protein}. Superfamily set (\texttt{SABMARK-sup}) contains 425 groups of multiple alignments, while the twilight zone set (\texttt{SABMARK-twi}) contains 209 groups.
    
    \item \underline{SCOP} \cite{andreeva2020scop} (\url{https://scop.berkeley.edu}) database (v2.07) was used to derive a set of 59,092 unique protein domain pairs, randomly sampled from the superfamily (36,372) and family (22,720) levels of its hierarchy. These 59,092 pairs were aligned separately using DALI \cite{DALI} and MMLigner \cite{collier2017statistical} to provide \texttt{SCOP1} and \texttt{SCOP2} benchmark alignments.
    
\end{enumerate}
New stochastic models of amino acid exchanges are automatically inferred on the above benchmarks and performance compared to popularly used substitution matrices in Shannon information terms, without the necessity of hand-tuning parameters (as demonstrated in \mySection{sec:results_msglens}).

\subsection{Stochastic matrix $\submat{}$ to model amino acids in the matched regions}
\label{sec:stochasticMatrixIntro}

Amino acid interchanges are modelled here by a Markov chain \cite{norris1998markov} defined over the state space of 20 amino acids. The probabilities of transitions between any pairs of amino acid states is represented here as a stochastic matrix $\submat{}$.
For any discrete time interval $t>0$, if an amino acid $a_0$ (at time $t=0$) undergoes the following chain of interchanges ($a_0 \rightarrow a_1 \rightarrow ... \rightarrow a_{t-1} \rightarrow a_t$), the Markov process ensures that the state of the amino acid $a_t$ (at time $t$) depends only on the previous state of the amino acid $a_{t-1}$ (at time $t-1$): 
\begin{align*}
    \prob{a_t|a_0,a_1,...,a_{t-1}} = \prob{a_t|a_{t-1}}.  
\end{align*}
Thus, the conditional probability $\prob{a_t |a_{t-1}}$ corresponds to a single step transition between the two states as observed after one discrete unit of time from the time $t-1$.

In this work, $\submat{}$ is represented by a $20\times 20$ matrix containing conditional probabilities, where each cell $\submat{ij}$ gives to the probability of an amino acid indexed by $j$ changing to an amino acid indexed by $i$, in one time-unit (note, $0 \leq \submat{ij} \leq 1$ ). Although this time-unit can be arbitrarily defined, it should be small and we use the convention introduced by Dayhoff et al. (1978) \cite{dayhoff1978}, and define one time-unit as the time taken for observing a 1\% (= 0.01) \emph{expected} change in amino acids under the model defined by the matrix. 

We term the probability matrix $\submat{}$ at time $t=1$, i.e. $\submat{} = \mathbf{M}(1)$, the \emph{base matrix}. In our formulation, each column vector of $\submat{}$ is an $\mathbb{L}_1$ normalised vector (i.e. 
$\sum_{i=1}^{20} \submat{ij} =1$ for all $1 \leq j \leq 20$). Further, as the Markov property holds, $\submat{}(t)$ can be computed from $\submat{}(1)$ as $\submat{}^t$, denoting the stochastic matrix after t time-steps. Implicit in $\submat{}$ is its stationary distribution $\pi$ \cite{norris1998markov} such that 
$\lim_{t \rightarrow \infty} \submat{}^t$ gives a matrix whose columns all tend to $\pi$. This stationary distribution $\pi$ is derived from the eigenvector corresponding to the eigenvalue of $1$ (the largest eigenvalue) of $\submat{}$.

\subsection{Multinomial probabilities $\nullmodel$ to model the amino acids in the indel region} 
\label{sec:indelModelIntro}

In general, the multinomial probabilities can be estimated over observations from any finite alphabet $\alphabet=\set{x_1,...,x_{|\alphabet|}}$ with $|\alphabet|$
symbols/states.  The \citet{wallace1987estimation} MML-estimate for multinomial probabilities has been derived as (refer 
\cite{allison2018coding}): 
\begin{equation*}
    \prob{x_i} = \frac{n_i + \frac{1}{2}}{\sum_{j=1}^{|\alphabet|}n_j + \frac{|\alphabet|}{2}}
\end{equation*}
where $n_i$ is the number of observations of each state $x_i$.  Using the above, we derive the optimal MML probability estimates $\nullmodel$ of each amino acid by accounting for the number of observations of each amino acid in the indel regions of all alignments in any specified collection $\data$.

To provide an alternate estimate for $\nullmodel$, we compare the above optimal choice with those derived from the stationary distribution $\pi$ of the stochastic matrix $\submat{}$. We note that $\pi$ is the eigenvector corresponding to $\submat{}$’s largest eigenvalue whose value is one.  Further, both the above candidates for estimates of $\nullmodel$ are dependent on the observed data $\data$. To compare these against an estimate independent of $\data$, $\nullmodel$ can be estimated on the UniProt database \cite{apweiler2004uniprot}
(see \SMSection{2.2}).

\subsection{Alignment 3-state machine and Dirichlet distributions as a function of time}
\label{sec:dirichletIntro} \label{sec:dirichletIntro}

\begin{figure}[!hbt]
    \centering
    \includegraphics[scale=0.2]{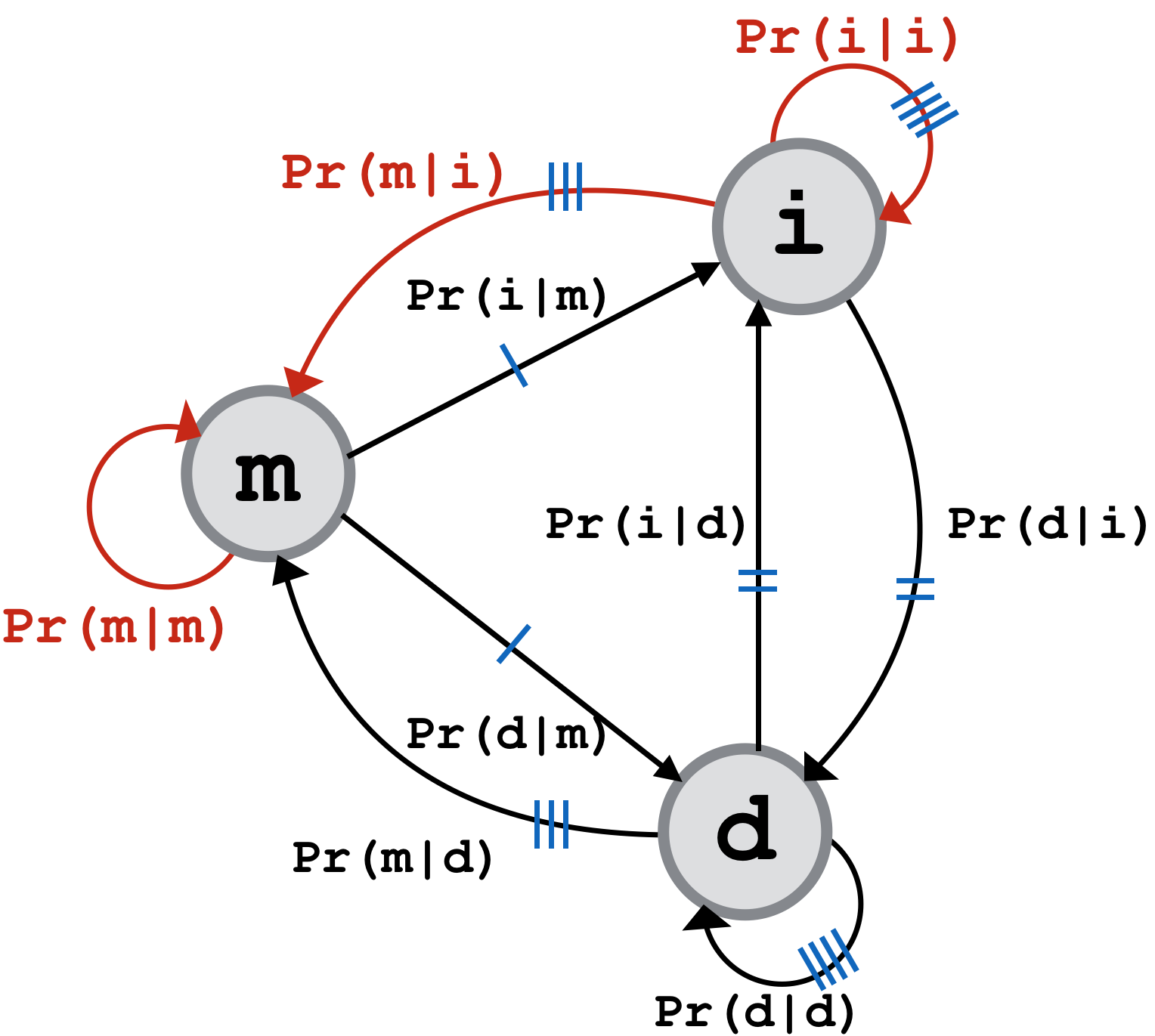}
    \caption{The symmetric 3-state machine for modelling an alignment string (Note: the red colour highlighted transitions refer to the three free parameters. Equivalent transitions are indicated in blue signs)}
    \label{fig:fsm1}
\end{figure}

In the collection $\data$, any alignment relationship $\alignment{i}$ over its corresponding sequences $\seqpair{i}$ is described as a 3-state string over the alphabet of \{\match\ (\Mstate), \insertion\ (\Istate), \delete\ (\Dstate)\} states, as previously considered by \citet{gotoh1982improved} and \citet{allison1992finite}. 
This 3-state machine defines nine possible one-step state transitions, with corresponding transition probabilities (denoted as $\stateParams$) (see \myFig{fig:fsm1}), where the sum of probabilities out of any state equals 1. Further, as is common when dealing with alignments of biomolecules, the \insertion\ and \delete\ states are treated symmetrically, thus reducing the number of  free parameters to three. Notionally these free parameters are represented by
$\set{\Pmm,\Pii,\Pmi}$. The remaining six (dependent) parameters can be can be derived from these as:
$\Pim = \Pdm = \frac{1-\Pmm}{2}$; $\Pdi = \Pid = 1 - \Pii - \Pmi$; $\Pdd = \Pii$; $\Pmd=\Pmi$.

These 3-state machine parameters are modelled using inferred Dirichlet distributions for different values of time $t$. In general, a Dirichlet distribution is a natural choice to model the parameters of multistate (categorical) and multinomial
distributions as the Dirichlet is a conjugate prior of the latter.
The time-dependent Dirichlet parameters inferred from $\data$ are denoted as $\dirparamst{t}$.
See \SMSection{1.3.4} for the methodological details of the inference of time-dependent Dirichlet parameters $\dirparamst{t}$,  and $\stateParams$ given $\dirparamst{t}$,  from any collection of alignments $\data$.

\subsection{Converting a substitution scoring matrix to a stochastic matrix $\submat{}$}
\label{sec:matrixConversion}

Any existing substitution matrix can be converted into its corresponding stochastic matrix M and thus benefit from the MML’s unsupervised parameter estimation. This lends the ability to objectively compare the performance of commonly-used scoring matrices over amino acid substitutions. These scoring matrices are normally published in their linearly-scaled \emph{log-odds} form. These log-odds matrices are converted back to their conditional probability form, by using their reported multiplier and amino acid frequencies.\footnote{Two (MIQS and PFASUM) of the nine substitution matrices we compared do not provide the amino acid frequencies used in their computation of log-odds scores, so we used the multinomial probability estimates of amino acids optimal for the collection $\data$} Let $\conditionalmat$ denote the conditional probability matrix derived from a substitution scoring matrix. Then, $\submat{}$ is derived from $\conditionalmat$ by numerically identifying the $k$-th root of the matrix $\conditionalmat$, i.e. 
$\submat{}(1) = \conditionalmat^{\frac{1}{k}}$, such that the resulting $\submat{}$ is nearest to 1\% (=0.01) expected amino-acid change (refer \SMSection{1.5.2}).

Once the stochastic matrix is derived from an existing substitution scoring matrix, all the other corresponding parameters ($\nullmodel$, $\stateParams$, $\dirparams$, $\boldsymbol\tau$) are automatically inferred, using MML, to be optimal to $\submat{}$ for the given collection $\data$.

\subsection{Inference of the best stochastic matrix $\submat{}^*$ for any benchmark $\data$}
\label{sec:inferBestMatrix}

This MML framework also allows the inference of the stochastic matrix from $\data$, the  best matrix being the one that minimises the MML objective function given in \myEqn{eq:2}. To search for the best stochastic matrix over any $\data$, we implement a Monte Carlo search method (see \SMSection{1.4}). Broadly, beginning from an initial state of $\submat{}$, randomly chosen columns of the evolving matrix are perturbed in the near-neighborhood. Using the Metropolis criteria each perturbation is either accepted/rejected over an iterative Monte Carlo process until convergence (refer \SMSection{1.4}).  

\subsection{ Estimation of terms in \myEqn{eq:2}} 
\label{sec:mmlTermsEstimation}

The `MML87' method \cite{wallace1987estimation} of parameter estimation is used to compute the Shannon information terms in \myEqn{eq:2}. In general, for a model with continuous parameters $\eta$, and prior $h(\eta)$, Wallace and Freeman (1987) derive the following message length required to explain any observed data-set $d$:
\begin{align*}
    \I{\eta,d} = \I{\eta} + \I{d|\eta}
\end{align*}
where
\begin{equation*}
    \I{\eta} = -\log{[h(\eta)]} + \frac{1}{2} \log{\{det[Fisher(\eta)]\}} + \frac{|\eta|}{2} \log{(\kappa_{|\eta|})} + \frac{|\eta|}{2}
\end{equation*}
($|\eta|$ is the number of free parameters, and $\kappa_{|\eta|}$ is
the associated lattice constant \cite{conway1984}). 

\begin{equation*}
    \I{d|\eta} = \mathcal{L}(d) 
\end{equation*}
$det(Fisher(\eta))$ is the determinant of the Fisher information matrix (the second derivative of the negative log likelihood function) that informs the optimal precision required to state $\eta$.
Individual statement lengths of terms in \myEqn{eq:2} are described in  the \SMSection{2}.

\bibliographystyle{abbrvnat}
\bibliography{references}

\end{document}

\title{Supplementary notes for ``\textit{Bridging the Gaps in Statistical Models of Protein Alignment}''}
\author{%
\name{Dinithi Sumanaweera, Lloyd Allison$^*$  and Arun Konagurthu\thanks{Corresponding Author}}
\address{Department of Data Science and Artificial Intelligence, Faculty of IT, Monash University, Australia}
}

\maketitle

\section{Supplementary Methods}
\small
\subsection{Overview of the inference methodology}
\label{sec:emoverview}
As described in Section 3.2 of the main text, given any benchmark $\mathbf{D}$ containing a collection of alignments (as 3-state strings) and their associated protein sequences,  our framework involves the simultaneous inference of the following statistical models:
\begin{itemize}
    \item the stochastic matrix $\mathbf{M}$ used to explain the matched pairs of amino acids observed in $\mathbf{D}$;
    \item the multinomial probability estimates $\mathbf{P}$ to explain the amino acids in the indel regions in $\mathbf{D}$;
    \item the divergence time parameter $t_i\in\boldsymbol{\tau}$, one for each alignment in $\mathbf{D}$;
    \item the 3-state transition probabilities $\mathbf{\Theta}_i$ for each alignment 3-state string in $\mathbf{D}$; and
    \item the ``time''-dependent Dirichlet distributions $\boldsymbol\alpha$ modelling the 3-state probabilities of all alignments in $\mathbf{D}$.
\end{itemize}

Of the above, the inference of the optimal MML estimates of multinomial probabilities $\mathbf{P}$ is fully independent of the inference of other models. These MML estimates are dependent solely on the insertions and deletions observed in the benchmark $\mathbf{D}$. However, the remaining parameters $\mathbf{M}, \boldsymbol\alpha, \mathbf{\Theta}$, and $\boldsymbol{\tau}$ have to be inferred simultaneously. Therefore, we rely on an iterative approach similar to an Expectation-Maximisation (EM) strategy. Under a monotonically decreasing value of the objective function (in this case the lossless encoding length), a set of parameters are held fixed (specifically $\boldsymbol\alpha$ and $\mathbf{\Theta}$) to optimise for the remaining parameters (specifically, $\mathbf{M}$ and $\boldsymbol\tau$). Then these ($\mathbf{M}$ and $\boldsymbol\tau$) are held fixed, to optimise the former ($\boldsymbol\alpha$ and $\mathbf{\Theta}$). This process is repeated until the total lossless encoding length does not change between successive iterations.

Broadly, the method of \citet{wallace1987estimation} (referred to as the MML87 method) lays the foundation on which model-parameters are inferred, and respective encoding lengths (given in Eq. 2 of the main text) are estimated. As stated in section 3.9 (main text), for any model with continuous parameters $\eta$, and prior $h(\eta)$, \citet{wallace1987estimation} derived the following message length required to explain any observed data-set $d$:
$
    I(\eta,d) = I(\eta) + I(d|\eta)
$
with
\newcommand{\conway}{c}
\begin{eqnarray}
\label{eqn:mml87firstpart}
    I(\eta) &=& -\log{[h(\eta)]} + \frac{1}{2} \log{\{det[Fisher(\eta)]\}} + \frac{|\eta|}{2} \log{(\conway_{|\eta|})} + \frac{|\eta|}{2}\\
\label{eqn:mml87secondpart}
    I(d|\eta) &=& \mathcal{L}(d) 
\end{eqnarray}
where $|\eta|$ is the number of free parameters; $\conway_{|\eta|}$ is
the associated lattice constant \cite{conway1984}; and  $det(Fisher(\eta))$ is the determinant of the \emph{expected} Fisher information matrix (the expected value of the second derivative of the negative log likelihood function) that informs the optimal precision required to state $\eta$. In MML, only the free parameters require encoding (since all dependent parameters can be deduced from the free ones).

Below, we will first discuss the estimation of encoding lengths of various terms involved in Eq. 2  in the main text, followed by the details of the search methods for optimising the aforementioned model-parameters.

% bookmark sec.1.2  p.1
\subsection{Introduction to Dirichlet distributions}
Dirichlet distributions are used here to losslessly encode (using MML) the transition probabilities of a 3-state machine. In general, a Dirichlet probability distribution models $d$-dimensional points in the standard unit $(d-1)$-simplex~\cite{allison2018codingChapter9}. 
%\cite[chapter 9]{allison2018codingChapter9}).
We denote a Dirichlet distribution as $\mathrm{Dir}(\vec{\alpha})$,
with a parameter vector \dirParams{} = \dirParamVec{}. This models the distribution of a random variable \dirData{}  = \dirDataVec{} (such that $\sum_{i=1}^{\text{\dirDim}}$\dirDatum{i} = 1) 
% LA3: see above
whose values represent a $d$-dimensional probability vector. 
We note that \dirData{} has $d-1$ degrees of freedom, while \dirParams{} has $d$  degrees of freedom.

Dirichlet distributions are best understood using the following reparameterisation of \dirParams{}: 
\label{sec:dirReparam}
\begin{center}\label{eq:reparam} 
\dirParams{} = \dirConcentration{}$\cdot$\dirMean{}
=
$\underbrace{\text{\dirConcentrationExpanded{}}}_{\text{\dirConcentration}}\cdot \underbrace{\text{\dirMeanExpanded{}}}_{\text{\dirMean}}$
\end{center}
where the distribution's spread is controlled by the concentration parameter $\kappa$ defined about the distribution's mean $\mathbb{L}1$-normalized vector $\hat{\mu}$. In this work,  all ``time''-dependent \dirParams\ are inferred in terms of their respective $\kappa$ and $\hat{\mu}$ values.

A Dirichlet probability density function is denoted as:
\begin{equation} \label{eq:dirPDF}
\mathrm{Dir}(\vec{\Theta};\vec{\alpha}) = \frac{1}{\text{\betaFunction{\dirParams{}}}} \displaystyle \prod_{i=1}^{\text{\dirDim}}\text{\dirDatum{i}}^{\text{\dirParam{i}}-1} \quad 
\end{equation}
where  
$\displaystyle
    \text{B(\dirParams{})} =  \frac{\prod_{i=1}^{\dirDim} \Gamma(\alpha_i)  }{\Gamma(\kappa)}
$  
is the multivariate Beta function defined using the Gamma function ($\Gamma(.)$)

\paragraph{Computing the Fisher information of Dirichlet distribution.}

The Fisher information matrix of Dirichlet distribution is given by the second derivative of its negative log likelihood function. The likelihood over 
% LA3: does 'observed' suggest it is not 'expected'??      observed 
data $\mathbf{\Theta}$ containing \dirDataCount{} samples $\mathbf{\Theta}=\{\vec{\Theta}_1,\ldots,\vec{\Theta}_N\}$ (where each $\vec{\Theta}_j, \forall 1\le j\le N,$ is of the form $[\theta_{j_1},\ldots,\theta_{j_d}]$), is defined by the function $f(\mathbf{\Theta}|\text{\dirParams}) =  \prod_{j=1}^{\text{\dirDataCount}}$ Dir(\dirData{j};\dirParams); the negative log likelihood function is of the form:
$$
\mathcal{L}(\mathbf{\Theta}|\vec{\alpha}) = -\text{\dirDataCount}\log{\text{\gammaFunction{\dirConcentration}}} +
       \text{\dirDataCount}\sum_{j=1}^{\text{\dirDim}}\log{\text{\gammaFunction{\dirParam{j}}}} -
	\sum_{i=1}^{\text{\dirDataCount}} \sum_{j=1}^{\text{\dirDim}}(\text{\dirParam{j}}-1)\log{(\text{\dirDatum{j_i}})}
$$
The matrix containing second partial derivatives of the above function with respect to the Dirichlet parameters \dirParams\ gives the Fisher information matrix, whose derivative is 
(refer \cite[section 2.3]{allison2018codingChapter9}) 
%\cite[Chapter 9, section 2.3]{allison2018coding}):
\begin{align} \label{eq:dirFisher}
det[Fisher(\text{\dirParams{}})] = \text{\dirDataCount}^{\text{\dirDim}}
					\{ \prod_{i=1}^{\text{\dirDim}}\text{\polygamma{1}{\dirParam{i}}} \}	
					\{ 1 - \text{\polygamma{1}{\dirConcentration}} (\sum_{i=1}^{\text{\dirDim}}  
					\frac{1}{\text{\polygamma{1}{\text{\dirParam{i}}}}}  )  \}
\end{align}
where \polygamma{1}{$\cdot$} is the polygamma function of order $1$ (a.k.a. trigamma function). Note, the determinant of the associated Fisher matrix, $Fisher(\vec{\alpha})$, indicates how sensitive the expected negative log likelihood function $\mathcal{L}$ is to the changes of \dirParams{}. This determinant is among the terms that dictates the optimum precision to which continuous parameters (in this case \dirParams) need to be stated, and influences their optimal encoding length (see Eq. \ref{eqn:mml87firstpart}).

% LA3: ?? do you happen to know if the sampling method has a "name" ??
% bookmark  p.2
\paragraph{Sampling a $\dirDim$-dimensional probability vector from a given Dirichlet distribution.}
\label{sec:dirSampling}

A $\dirDim$-dimensional probability vector $\vec{\Theta} = (\theta_1,...,\theta_\dirDim)$ can be sampled from a $d$-dimensional $\mathrm{Dir}(\vec{\alpha})$ as follows:
\begin{enumerate}
    \item Each component $\theta_i$ is treated as a Gamma distributed sample $y_i$ generated from Gamma distribution $\mathrm{Gamma}(\alpha_i,1)$. 
    \item Repeating the above $\dirDim$ times, we get a vector $\vec{y} = (y_1,\ldots,y_d)$
    \item $\vec{y}$ is $\mathbb{L}_1$-normalized to generate $\vec{\Theta}$.
\end{enumerate}
The resultant vector $\vec{\Theta}$ with $\theta_{i}=\frac{y_i}{\sum_{i=1}^{\dirDim}y_i}$ is Dirichlet distributed under $\mathrm{Dir}(\vec{\alpha})$. 

\paragraph{Message length of stating a probability vector $\vec{\Theta}$ using a Dirichlet model $\mathrm{Dir}(\vec{\alpha})$.}
In general, let  $\vec{\Theta} = (\theta_1,...,\theta_\dirDim)$ denote a $d$-dimensional probability vector, where each $\theta_i$ denotes the probability of each distinct state $x_i$, whereas $count(x_i)$ is the number of times $x_i$ is observed. This implies, $\vec{\Theta}$ has $(d-1)$ degrees of freedom  (notionally, say $\theta_d=1-\sum_{i=1}^{d-1} \theta_i$).  Since, $\sum_{i=1}^d \theta_i=1$, $\vec{\Theta}$ is a point inside a $(d-1)$-simplex, modelled using a Dirichlet distribution $\mathrm{Dir}(\vec{\alpha})$.

Using the MML method of \citet{wallace1987estimation}, the statement cost of $\vec{\alpha}$ and $\vec{\Theta}$ given $\vec{\alpha}$ (denoted by $\vec{\Theta}|\vec{\alpha}$) is given by:
$
I(\vec{\alpha},\vec{\Theta}) = I(\vec{\alpha}) + I(\vec{\Theta}|\vec{\alpha})
$
where,
\begin{eqnarray}
%\begin{align} 
\label{eqn:IAlphaGeneral}
 \displaystyle
 I(\text{\dirParams{}}) &=& - \log{[h(\vec{\alpha})]} + \frac{1}{2}\log{(det[Fisher(\vec{\alpha})])} + \frac{d}{2}\log{(c_{d})} + \frac{d}{2} \\
 \label{eqn:IthetaGalphaGeneral}
 \displaystyle
      I(\vec{\Theta}|\vec{\alpha}) &=&  - \log{[\mathrm{Dir}(\vec{\Theta}_{i};\text{\dirParams{}})]} + \frac{1}{2}\log{(det[Fisher(\vec{\Theta}_{i})])} +\frac{d-1}{2}\log{(c_{d-1})} +  \frac{d-1}{2} 
\end{eqnarray}
In Equations \ref{eqn:IAlphaGeneral}-\ref{eqn:IthetaGalphaGeneral}, $h(\text{\dirParams{}})$ is any specified prior on Dirichlet parameters \dirParams{}, whereas
$\mathrm{Dir}(\text{\dirParams{}})$ is the Dirichlet prior on $\vec{\Theta}$. \latticeConst{d} and \latticeConst{d-1} are the optimal lattice constants~\cite{conway1984} associated with $d$ degrees of freedom in $\vec{\alpha}$ and $(d-1)$ degrees of freedom in $\vec{\Theta}$, respectively. Further, $det[Fisher(\text{\dirParams{}})]$ is the determinant of the expected Fisher information of \dirParams{} given in Equation \ref{eq:dirFisher}, that informs the optimal precision to which that parameter should be stated in MML. Similarly, $det[Fisher(\vec{\Theta})]$ given in Equation \ref{eq:thetaFisher} below is the expected Fisher information of $\vec{\Theta}$ that informs the optimal precision to which $\vec{\Theta}$ is stated, as computed by MML \cite{boulton1969information,wallace1987estimation}.
\begin{align}
\label{eq:thetaFisher}
    det[Fisher(\vec{\Theta})] = \frac{\left(\sum_{i=1}^d count(x_i)\right)^{d-1}} %{n(\text{\stateX{}})}^{k-1}}
    { \displaystyle \theta_{i}\theta_2\cdots\theta_d}
\end{align}
Note, $\sum_{i=1}^d count(x_i)$ is the total number of observations (combining all states).

% bookmark  sec.1.3  p.3
\subsection{Computing the terms in Equation 2 of the main text}
\subsubsection{Computation of $I(\mathbf{M})$}
A stochastic matrix $\mathbf{M}$ is a $20\times 20$ matrix where the value in any cell $M_{ij}$  is the conditional probability of an amino acid indexed by $j$ changing to an amino acid indexed by $i$, in one unit of time. Since $\sum_{i=1}^{20}\mathbf{M}_{ij}=1$, each column $M_j$ is a $(d=20)$-dimensional $\mathbb{L}_1$-normalized multinomial probability vector (over the amino acid states). Therefore, the statement cost of the whole matrix $\mathbf{M}$ can be described as the sum of individual statement costs associated with each of its 20 columns:
$$
I(\mathbf{M}) = \sum_{j=1}^{20} I(\mathbf{M}_j)
$$
Using a uniform prior on each $\mathbf{M}_j$ (which implies a Dirichlet distribution with $\vec{\alpha}=\{1,1,\ldots,1\}$), the statement lengths  $I(\mathbf{M}_j), \forall 1\le j\le 20$ can be computed using Equation \ref{eqn:IthetaGalphaGeneral}.

% bookmark  sec.1.3.2  p.3
\subsubsection{Computation of $I(\mathbf{P})$}

In general, the MML87 estimate $\theta_i\in \vec{\Theta}$ for any state $x_i$ that is observed $counts(x_i)$ times, is derived as:
% LA3: should be wallace and boulton 1968 (or 1967) if anyone  (see \cite[chapter 9]{allison2018coding}):
\begin{align} \label{eq:mmldirichletmultistate}
    \theta_i = \frac{counts(x_i) +\alpha_i - \frac{1}{2}}{\sum_{i=1}^{d} (counts(x_i) + \alpha_i) - \frac{d}{2}} 
\end{align} 
Here, $\mathbf{P}$, that is used in this framework to encode amino acids in the indel regions of alignments in a given benchmark $\mathbf{D}$, is therefore inferred as a $(d=20)$-nomial probability vector containing the probability estimates of all distinct amino acids, as observed in the indel regions. Thus, the 20-dimensional $\mathbf{P}$ vector can be inferred from observations of indels in $\mathbf{D}$, as per Equation \ref{eq:mmldirichletmultistate}.

Further,  under the assumption that the prior on $\mathbf{P}$ is  uniform (i.e. $\vec{\alpha}=\{1,1,\ldots,1\}$), the statement length  $I(\mathbf{P})$ can be computed again using Equation \ref{eqn:IthetaGalphaGeneral}.

\subsubsection{Estimating time $t_i\in\boldsymbol\tau$ of any alignment $\mathcal{A}_{i}\in\mathbf{D}$ and the computation of $I(t_i)$}

For a given set of parameters $\mathbf{M}, \mathbf{P}, \boldsymbol\alpha, \mathbf{\Theta}$ (i.e. when they are held fixed within any iteration of our EM-like approach -- refer Section \ref{sec:emoverview}), these parameter values allow the estimation of an optimal each time parameter $t_i\in\boldsymbol\tau$ (modelling the divergence of each sequence-pair in $\mathbf{D}$) that is optimal to those parameters, as follows.

Using each $\left<\mathcal{A}_i,\mathbf{S}_i,\mathbf{T}_i\right>\in \mathbf{D}$, we want to estimate the best (integer) value of $t_i$ for a fixed stochastic matrix \submat{} (i.e. $\mathbf{M}(t_i) = {\mathbf{M}}^{t_i}$ ) and associated parameters stated above, such that the total message length of stating \alignment{i}, \seqS{i} and \seqT{i} is minimised: 
\begin{eqnarray*} \displaystyle
I(\mathcal{A}_{i},\left<\text{\seqS{i},\seqT{i}}\right>|\text{\submat{t},} \boldsymbol\alpha, \mathbf{\Theta} ) =  I(\mathcal{A}_{i}|\boldsymbol\alpha, \mathbf{\Theta}, t_i) 
~~+
&\sum_{\substack{
\forall 
\mathcal{A}_i(j) = \mathbf{match}\\
\& \mathcal{A}_i(j)
\implies 
\left(\substack{
\text{\seqS{i}}(k)\\
\text{\seqT{i}}(l)
}\right)
}}
&
I(\text{\seqS{i}}(k),\text{\seqT{i}}(l)|\mathbf{M}(t_i)) \\\displaystyle
+
&\sum_{\substack{
\forall 
\mathcal{A}_i(j) = \mathbf{delete}\\
\& \mathcal{A}_i(j)
\implies 
\left(\substack{
\text{\seqS{i}}(k)\\
\text{-- (gap)}
}\right)
}}
&
I(\text{\seqS{i}}(k)|\mathbf{P}) \\
+
&
\sum_{\substack{
\forall 
\mathcal{A}_i(j) = \mathbf{insert}\\
\& \mathcal{A}_i(j)
\implies 
\left(\substack{
\text{-- (gap)}\\
\text{\seqT{i}}(l)
}\right)
}}
&
I(\text{\seqT{i}}(l)|\mathbf{P}) 
\end{eqnarray*}
where $\mathcal{A}_i(j)$ is the $j$-th state in the 3-state string $\mathcal{A}_i$, $\mathbf{S}_i(k)$ is the $k$-th amino acid in the sequence $\mathbf{S}_i$, and similarly for $\mathbf{T}_i(l)$.

Using the above objective function, %($Obj$), 
we implemented a variant of the bisection method  % <-amen
over the integral values of $t\in[1,1000]$, where in each iteration, the interval is reduced by half (i.e. the search range is halved).
The variation attempts to handle some special cases to avoid the method from being trapped within in a local optima.

Finally, once $t_i$ is inferred for $\left<\mathcal{A}_i,\mathbf{S}_i, \mathbf{T}_i\right>$, optimal to the parameters $\mathbf{M}, \mathbf{P}, \boldsymbol\alpha, \mathbf{\Theta}$, the statement cost of integer $t_i$ uniform in the range $[1,1000]$ takes $\log_2(1000)$ bits to encode.

% bookmark  p.4  sec.1.3.4   24/9/2020
%
\subsubsection{Computation of  $I(\boldsymbol\alpha)$,  $I(\mathbf{\Theta}_i|\boldsymbol\alpha, t_i)$, and $ I(\mathcal{A}_i|\mathbf{\Theta}_i,\boldsymbol\alpha, t_i)$
%, and $I(\mathbf{A}|\mathbf{\Theta},\boldsymbol\alpha)$ 
}

The computation of these terms are feasible when the time parameters $\boldsymbol\tau=\{t_1,\ldots,t_{|\mathbf{D}|}\}$ for each aligned sequence-pair $\left<\mathcal{A}_i,\mathbf{S}_i, \mathbf{T}_i\right>\in\mathbf{D}$ are known. We first discuss the relationship between Dirichlet and three-state machine models, before exploring the computation of their respective message length terms.

\paragraph{Relationship between Dirichlet parameters and 3-state machine transition probabilities} 
Section 3.6 in the main text introduces the alignment 3-state machine over \texttt{match} (\match), \texttt{insert} (\insertion), and \texttt{delete} (\deletion) states, with 9 transition probability parameters between any two states. As shown in main text's Figure 4, the alignment 3-state machine is symmetric between \texttt{insert} (\insertion) and \texttt{delete} (\deletion) states, thus yielding 3 free parameters (and remaining 6 dependent). 

Notionally, the \texttt{match} state has only 1 free parameter, $\Pr(\match|\match)$.  The \texttt{insert} state has 2 free parameters, $\Pr(\insertion|\insertion)$ and $\Pr(\match|\insertion)$. The \texttt{delete} state is treated as symmetric to the \texttt{insert} state. Thus, remaining 6 transition probabilities can be derived from the symmetry between \texttt{insert} and \texttt{delete} states, and from the constraint that all transition probabilities out of any state in the machine adds up to 1.

All 9 transition probabilities in a alignment 3-state machine can be derived from the following composition of $\mathbb{L}_1$-normalized vectors, $\vec{\Theta}^{(\mathbf{match})}$ (defining a point in 1-simplex) and $\vec{\Theta}^{(\mathbf{insert})}$ (defining a point in 2-simplex):
\begin{eqnarray*}
\vec{\Theta}^{(\mathbf{match})} 
&=& \left\{\underbrace{\Pr(\match|\match)}_{\text{free}}, \underbrace{1-\Pr(\match|\match)}_{\text{dependent}}\right\}\\
\vec{\Theta}^{(\mathbf{insert})} 
&=& \left\{
\underbrace{\Pr(\insertion|\insertion)}_{\text{free}}, 
\underbrace{\Pr(\match|\insertion)}_{\text{free}}, 
\underbrace{1-\Pr(\insertion|\insertion)-\Pr(\match|\insertion)}_{\text{dependent}}\right\} 
\equiv
\vec{\Theta}^{(\mathbf{delete})} ~~\text{(by symmetry)} 
\end{eqnarray*}
Therefore, 
$\vec{\Theta}^{\mathbf{(match)}}$ is a variable modelled using a 1-simplex Dirichlet distribution \dirModelM, with \dirParamsM\ having $d=2$ degrees of freedom, and $\vec{\Theta}^{\mathbf{(insert)}}$ is a variable modelled using a 2-simplex Dirichlet distribution \dirModelI, with \dirParamsI\ having $d=3$ degrees of freedom. The MML estimates for $\vec{\Theta}^{\mathbf{(match)}}$ and $\vec{\Theta}^{\mathbf{(insert)}}$ is as per the general case given in Equation \ref{eq:mmldirichletmultistate}.

To infer time-dependent Dirichlet distributions, we partition the alignments in the benchmark $\mathbf{D}$ into time ($t$) related integer bins in the range $t\in[1,1000]$. The subset of alignments in each time-bin is denoted as  $\mathbf{A}(t)$, where $\mathbf{A}(t)\subset\mathbf{D}$. 

Given each time-bin with the subset of alignments $\mathbf{A}(t)$ containing $|\mathbf{A}(t)|$ number of alignment 3-state strings, the goal is to infer the time-dependent Dirichlet parameters $\boldsymbol\alpha(t) = \left\{\text{\dirParamsM},  \text{\dirParamsI}\right\}$ for that bin that model the observed set of 3-state machine parameters $\mathbf{\Theta}(t) =\left\{ \mathbf{\Theta}_i \equiv\left[\vec{\Theta_i}^{(\mathbf{match})}, \vec{\Theta_i}^{(\mathbf{insert})}\right]\right\}_{1\le i \le |\mathbf{A}(t)|}$ inferred for each alignment $\mathcal{A}_i\in\mathbf{A}(t)$.
 % ===================================================================

\paragraph{Message length terms}
%The objective is to infer 1-simplex and 2-simplex Dirichlet models (\dirParamsM and \dirParamsI) to model the free parameters $\vec{\Theta}^{\mathbf{(match)}}$ and $\vec{\Theta}^{\mathbf{(insert)}}$ of the 3-state machine, inferred on a set of 3-state strings \structAlignments. 

 %Thus, 
 We want to infer $\boldsymbol\alpha(t)$ and $\mathbf{\Theta}(t)$ using the observation of alignment states in $\mathbf{A}(t)$, by minimising the objective function:
\begin{equation}
 \label{eq:dirMMLTot}
I(\boldsymbol\alpha(t),\mathbf{\Theta}(t),\mathbf{A}(t)) 
= 
  I(\boldsymbol\alpha(t)) 
+ I(\mathbf{\Theta}(t)|\boldsymbol\alpha(t)) 
+ I(\mathbf{A}(t)|\mathbf{\Theta}(t),\boldsymbol\alpha(t))
\end{equation}
where the terms on the right-hand side can be further decomposed as:
\begin{eqnarray}
\label{eqn:componentAlpha}
I(\boldsymbol\alpha(t)) 
&=& 
I(\vec{\alpha}^{(\mathbf{match})}) 
+
I(\vec{\alpha}^{(\mathbf{insert})}) \\
\label{eqn:componentTheta}
I(\mathbf{\Theta}|\boldsymbol\alpha(t)) 
&=&
\sum_{i=1}^{|\mathbf{A}(t)|} 
\left(
I(\vec{\Theta}_i^{(\mathbf{match})}|\vec{\alpha}^{(\mathbf{match})},t_i=t)
~+~
I(\vec{\Theta}_i^{(\mathbf{insert})}|\vec{\alpha}^{(\mathbf{insert})},t_i=t)
\right)\\
\label{eqn:componentAlgn}
I(\mathbf{A}(t)|\mathbf{\Theta},\boldsymbol\alpha(t))
&=&
\sum_{i=1}^{|\mathbf{A}(t)|}
I(\mathcal{A}_i|\vec{\Theta}_i,\boldsymbol\alpha,t_i=t)
\end{eqnarray}

%%%%%IGNORE START
\ignore{
jointly, for any  $\mathbf{\left<state\right>}\in \{ \mathbf{match}, \mathbf{insert}\}$. 
Note: \structAlignmentsX{} contains only the data instances relevant to 
the state of focus. That is, for $\mathbf{\left<state\right>}=\mathbf{match}$, all state transitions of the form $\mathtt{match}\longrightarrow \{\mathtt{match, insert, delete}\}$ that are present in \structAlignments\ are considered.
Similarly, for $\mathbf{\left<state\right>}=\mathbf{insert}$, all state transitions of the form $\mathtt{insert}\longrightarrow \{\mathtt{match, insert, delete}\}$ that are present in \structAlignments\ are considered. The total two part message length is: 

\begin{align}  \label{eq:dirMMLTot}
\displaystyle
    I(\text{\dirParamsX{}, \ThetaX{}, \structAlignmentsX{}}) = 
    \underbrace{I(\text{\dirParamsX{}}) +
    I(\text{\ThetaX{} \given \dirParamsX{}})}_{\text{First part}} \qquad\\\nonumber+ \underbrace{I(\text{\structAlignmentsX{} \given \ThetaX{},\dirParamsX{} })}_{\text{Second part}} \quad \text{bits}
\end{align}
}
%%%%%%%IGNORE END

Further, each state-related Dirichlet parameters in the right-hand side of Eq. \ref{eqn:componentAlpha} of the form \dirParamsX{} where $\left<\mathbf{state}\right>\in\{\mathbf{match},\mathbf{insert}\}$, expands to the form (using method in \cite{wallace2005statistical}):

 \begin{align} \label{eq:Ialpha}
 \displaystyle
     I(\text{\dirParamsX{}}) =
     \frac{1}{2}\log{\bigg( 1 + \frac{det[Fisher(\vec{\alpha}^{\mathbf{(\left<state\right>)}})]  \times \left(\text{\latticeConst{d}}\right)^{d}}{ 
     h(\vec{\alpha}^{\mathbf{(\left<state\right>)}})^2  }   \bigg)}  +
     \frac{d}{2} 
 \end{align}
In the above expansion, $h(\text{\dirParamsX{}})$ is the prior on the state-related Dirichlet parameters \dirParamsX{}. \latticeConst{d} is the   % optimal 
lattice  % LA4: I think just 'lattice constant'  Conway-
constant \cite{conway1984} associated with $d$ degrees of freedom 
-- recall, $d=2$ for Dirichlet associated with the \texttt{match} state, % ?? <- one or two ??   ***
and $d=3$ for \texttt{insert} state;                                     % ?? <- two or three ?? ***
\latticeConst{2} = $\frac{5}{36\sqrt{3}}$ and 
\latticeConst{3} = $\frac{19}{192 \times 2^{\frac{1}{3}}}$.
Finally, $det[Fisher(\text{\dirParamsX{}})]$ is the determinant of the expected Fisher information of \dirParamsX{} given in Eq. \ref{eq:dirFisher}. For details of the prior used, see \citet[supplementary methods]{sumanaweera2019statistical}.

We note that the overall statement length of encoding all ``time''-dependent Dirichlet parameters for each time-bin ($\forall t\in[1,1000]$) is the sum of their individual bin-wise statement costs: $I(\boldsymbol{\alpha}) = \sum_{t=1}^{1000} I(\boldsymbol{\alpha}(t))$.

%%%%%%%IGNORE BLOCK
\ignore{
Further, using the reparameterization of a Dirichlet parameter \dirParams{} previously stated in Eq. \ref{eq:reparam} , the prior term can be decomposed as 
   $ h(\text{\dirParamsX{}}) = h(\text{\dirConcentrationX})h(\text{\dirMeanX})$, 
by applying independent priors for concentration parameter \dirConcentrationX{} and the $\mathrm{L1}$-normalized mean vector \dirMeanX{}. In this thesis, \dirMeanX{} is assumed to be
uniformly distributed in a unit $k-1$ simplex. Thus, $h(\text{\dirMeanX{}})$ is simply the reciprocal of the simplex volume. As a result, $h(\text{\dirParamsM{}}) = \frac{1}{\sqrt{2}}$ and $h(\text{\dirParamsI{}}) = \frac{2}{\sqrt{3}}$. 
On the other hand, \dirConcentrationX{} controls the Dirichlet concentration about the distribution's mode. For any $y$ diffusing with $k$ degrees of freedom, \cite{wallace2005statistical} defined a well-behaved prior for $y$, in the form:
\begin{align*}
    g(y) = \frac{y^{k-1}}{(1+y^{2})^{\frac{k+1}{2}}} 
\end{align*}
Before applying $g(\cdot)$ as a prior for \dirConcentrationX{}, a normalization constant is computed to ensure this as a valid probability distribution. This means: $\int_{0}^{\infty} g(y) = 1$. For $k=2$, we can directly take 
$h(\text{\dirConcentrationM{}}) = g(\text{\dirConcentrationM{}})$, as the above is satisfied. 
However for $k=3$, $\int_{0}^{\infty} g(y) = \frac{\pi}{4}$, which is not equal to 1. 
Therefore, this yields $h(\text{\dirConcentrationI{}}) = \frac{4}{\pi}g(\text{\dirConcentrationI{}})$. 
}
%%%%%%%IGNORE END
Similarly, each state-related transition probability parameters in the right-hand side of Eq. \ref{eqn:componentTheta} of the form $I(\text{\ThetaX{}\given \dirParamsX{}})$ (again $\left<\mathbf{state}\right>\in\{\mathbf{match},\mathbf{insert}\}$) expands to (using the methods in \cite{wallace2005statistical}):
 \begin{align} \label{eq:IthetaGalpha}
 \displaystyle
     I(\vec{\Theta}_{i}^{\mathbf{(\left<state\right>})}\text{\given\dirParamsX{}}) =  \sum_{i=1}^{|\mathbf{A}(t)|} 
     \left(
     \frac{1}{2}\log{\bigg( 1 + \frac{det[Fisher(\vec{\Theta}_{i}^{\mathbf{(\left<state\right>})})]  \left(\text{\latticeConst{d-1}}\right)^{d-1}}{
     (\mathrm{Dir}(\vec{\Theta}_{i}^{\mathbf{(\left<state\right>)}}; \text{\dirParamsX{}}))^2  }   \bigg)} + \frac{d-1}{2}\right)
 \end{align}
%Here, the Equation \ref{eq:dirPDF} provides the likelihood of \dirParamsX{} that describes $\vec{\Theta}_{n(\text{\stateX{}})}$. 
The determinant of the expected Fisher information of the inferred MML probability estimates $\vec{\Theta}^{\mathbf{(\left<state\right>)}}=(\theta_1,\theta_2,\ldots,\theta_d)$ (see Eq. \ref{eq:mmldirichletmultistate}), with $d-1$ free parameters (notionally, with dependent parameters $\theta_d= 1-\sum_{i=1}^{d-1} \theta_i$)  on each alignment $\mathcal{A}_i\in\mathbf{A}$ is given by:
\begin{align*}
    det[Fisher(\vec{\Theta}_{i}^{\mathbf{(\left<state\right>)}})] = \frac{\left(counts(\mathbf{\left<state\right>}\longrightarrow \{\mathtt{match, insert, delete}\})\right)^{d-1}} %{n(\text{\stateX{}})}^{k-1}}
    { \displaystyle \theta_{i}\theta_2\cdots\theta_d}
\end{align*}
where $counts(\mathbf{\left<state\right>}\longrightarrow \{\mathtt{match, insert, delete}\})$ is the number of observed transitions from the specified state to any of the $\{\mathtt{match, insert, delete}\}$ states in alignment \alignment{}$_i$.
Specifically, when $\mathbf{\left<state\right>} = \mathtt{match}$ we count the total number of \mmtransition{}, \mitransition{} and \mdtransition{} transitions in $\mathcal{A}_i$, while the denominator refers to the product of $d=2$ MML estimates for \prob{\match{\given\match{}}} and (1-\prob{\match{}\given\match{}}).
Similarly, when $\mathbf{\left<state\right>} = \mathtt{insert}$ (which is symmetric with \texttt{delete} state), we count the total number of \iitransition\ plus \ddtransition, \imtransition\ plus \dmtransition, \idtransition\ plus \ditransition{} transitions observed in $\mathcal{A}_i$, while the denominator refers to the product of $d=3$ MML estimates for  \prob{\insertion{}\given\insertion{}}, \prob{\match{}\given\insertion{}} and (1-\prob{\insertion{}\given\insertion{}}- \prob{\match{}\given\insertion{}} = \prob{\deletion{}\given\insertion{}}).

Finally, the statement of each 3-state string $\mathcal{A}_i$ in the time-bin $\mathbf{A}(t)$, given on the right-hand side of Eq.  \ref{eqn:componentAlgn}, can be expanded to:
 $$
 I(\mathcal{A}_i|\vec{\Theta}_i,\boldsymbol\alpha,t_i=t)
 =
 \sum_{j=1}^{|\mathcal{A}_i|} -\log(\Pr(\mathcal{A}_i(j)|\mathcal{A}_i(j-1))
 $$
 where $\mathcal{A}_i(j)$ is the $j$-th character in the 3-state string $\mathcal{A}_i$ whose length is $|\mathcal{A}_i|$, and $\Pr(.|.)$ is one of the 9 transition probabilities of the 3-state machine associated with the alignment $\mathcal{A}_i$, derived from $\vec{\Theta}_i$.
 
\small

% bookmark  p.5  sec.1.4   24/9/2020 -----------------------------------------------------------------------
%
\subsection{Monte Carlo search to infer all parameters over a given benchmark $\mathbf{D}$} 
\label{sec:dirOptimisationMCMC}
Using the various encoding length terms in Section 1.3,  we rely on an Expectation-Maximisation like strategy to simultaneously infer optimal parameters across all statistical models required to estimate the Shannon information content of any collection $\mathbf{D}$ (under those models).

Broadly, the approach starts from any specified initial values for $\mathbf{M}(t)$ and $\boldsymbol\alpha(t)$. A good starting point would be from any of the existing substitution matrices (after converting to their stochastic base ($t=1$) matrix form). 
% LA4: the following implies that Alpha is, or can be, compuer from a matric ??...
Similarly $\boldsymbol\alpha(t)$ inferred from any of the earlier matrices is a good starting point. The files containing our starting points are available here: \url{https://lcb.infotech.monash.edu.au/mmlsum}.

Note, as stated above in Section S1.3.2, the optimal MML estimates of $\mathbf{P}$ is solely dependent on the benchmark $\mathbf{D}$ and not on any other parameters. 

The EM-like search involves holding $\boldsymbol\alpha(t)$ and $\vec{\Theta}$ fixed while performing a Monte Carlo search to optimise for $\mathbf{M}$ and $\boldsymbol\tau$. (These details are presented in Section \ref{sec:matrixEstimation}) After this, $\mathbf{M}$ and $\boldsymbol\tau$ are held fixed to estimate $\boldsymbol\alpha(t)$ and $\vec{\Theta}$. (These details are presented in \ref{sec:dirichletEstimation}). This process is repeated until the objective function converges.

% bookmark  sec.1.4.1  p.5
\subsubsection{Search for time-dependent Dirichlet parameters with fixed values of $\mathbf{M}$ and $\boldsymbol\tau$} \label{sec:dirichletEstimation}

Holding the stochastic matrix $\mathbf{M}$ and time parameters $\boldsymbol\tau$ fixed, this allows us to partition all alignments in the benchmark $\mathbf{D}$ into respective time-bins, resulting in subsets of alignments $\mathbf{A}(t)$ for each $t\in[1,1000]$.

There is no closed-form to estimate the parameters of a Dirichlet distribution.  % LA4: <- added ??   (BTW there are some numerical-methods methods to do it.  Don't go there??)
For each $t\in[1,1000]$, starting from the current state of $\boldmath\alpha(t)$, we iteratively perform random perturbations on $\boldmath\alpha(t)$ in its near-neighborhood by either perturbing the mean vector or the concentration parameter (see pseudocode in Figure \ref{fig:algo}).

Specifically, in each iteration, there is 50-50 choice between \texttt{Perturb\_Mean} and \texttt{Perturb\_Concentration} functions (refer decomposition of Dirichlet into mean and concentration given in Section S1.2). The former perturbs the mean vector $\hat{\mu}$ of \dirParamsX{}  by sampling a new probability vector from a Dirichlet distribution of mean \dirParamsX{} and a specified concentration $\bar{\kappa}$. The latter randomly either increases or decreases the concentration parameter by $\pm\delta$ from its current value.
\ignore{
Here we choose $\bar{\kappa}=10000$, a higher concentration for state \match{}, compared to $\bar{\kappa}=1000$ for state \insertion{}. This is because, the 1-simplex model has less freedom due to one free parameter, making it more sensitive to a slight perturbation. On the other hand, 2-simplex model has two degrees of freedom, thus we impose more flexible perturbations. As  Dirichlet concentration decreases, the neighbourhood where a perturbed sample resides in the distribution expands. 
}

After each perturbation, the message length defined in Equation \ref{eq:dirMMLTot} is evaluated. If the message length decreases, it is accepted with a probability of 1, else the perturbation is accepted/rejected with a probability equivalent to 
$2^{-\Delta I}$, 
where $\Delta I$ is the observed difference in the message lengths (in bits).

At the end of this process, $\boldsymbol\alpha(t), \forall t\in[1,1000]$ are optimised for the fixed values of $\mathbf{M}$ and $\boldsymbol\tau$. $\mathbf{\Theta}$ are estimated from these new values of $\boldsymbol\alpha(t)$ as described in Section S1.3.4.

\begin{figure}[!t]
\noindent \fbox{
\begin{minipage}[t]{0.5\linewidth}
\vspace{0pt}
\begin{algorithm}[H]
 \SetKwFunction{FMain}{Perturb\_Dirichlet\_Param}
 \SetKwProg{Pn}{Function}{:}{\KwRet{\dirConcentration{}.\dirMean{}}}
 \Pn{\FMain{\dirParamsX{}}}
 {
 $\{\text{\dirConcentration{},\dirMean{}}\} \leftarrow \texttt{decompose}(\text{\dirParamsX{}})$\; 
 $u_1 \leftarrow \texttt{random\_uniform}(0,1)$\;
  \If{$u_1 \leq 0.5$}{ 
      $\hat{\mu} \leftarrow$ $\texttt{Perturb\_Mean}(\text{\dirMean{},$\mathbf{\left<state\right>}$})$\;
     } 
\Else{
       % \dirConcentration{} $\leftarrow$ $\texttt{Perturb\_Prob\_Vec}(\text{\dirMean{},$\mathbf{\left<state\right>}$})$\;
$\kappa \leftarrow \texttt{Perturb\_Concentration}(\kappa)$\;
    }
    %\dirParamsX{} $\leftarrow$ \dirConcentration{}.\dirMean{}\; 
   % \dirParamsX{} $\leftarrow$ \dirConcentration{}.\dirMean{}\; 
}
\end{algorithm}
\begin{algorithm}[H]

 \SetKwFunction{FMain}{Perturb\_Mean}
 \SetKwProg{Pn}{Function}{:}{\KwRet{$\mathtt{ sample\_dirichlet}(\bar{\kappa}.\text{\dirMean{}}$)}}
 \Pn{\FMain{\dirMean{},$\mathbf{\left<state\right>}$}}
{
\If{$\mathbf{\left<state\right>}=\mathrm{match}$}{
    $\bar{\kappa} \leftarrow 10000$\;
}
\Else{
    $\bar{\kappa} \leftarrow 1000$\;
}
%$\bar{\alpha} \leftarrow \bar{\kappa}.\text{\dirMean{}}$\;
%$\text{\dirParamsX{}} \leftarrow $\texttt{ random\_dirichlet}($\bar{\alpha}$)\;
}
\end{algorithm}
\end{minipage} % no space if you would like to put them side by side
\begin{minipage}[t]{0.5\linewidth}
\vspace{0pt}

\begin{algorithm}[H]
 \SetKwFunction{FMain}{Perturb\_Concentration}
 \SetKwProg{Pn}{Function}{:}{\KwRet{$\kappa$}}
 \Pn{\FMain{$\kappa$}}
 {

   $\delta \leftarrow \texttt{random\_uniform}(0.1,10)$\;
    $u_2 \leftarrow  \texttt{random\_uniform}(0,1)$\;
    \If{$u_2 \leq 0.5$}{
        \dirConcentration{} $\leftarrow$ \dirConcentration{} + $\delta$\;
    }
    \Else{
        \dirConcentration{} $\leftarrow$ \dirConcentration{} - $\delta$\;
    }
 }
\end{algorithm}

\begin{algorithm}[H]

 \SetKwFunction{FMain}{sample\_dirichlet}
 \SetKwProg{Pn}{Function}{:}{\KwRet{$\vec{y}$}}
 \Pn{\FMain{$\vec{\alpha}$}}
{
 Init $\vec{y}\leftarrow \vec{0}, \textrm{sum} \leftarrow 0$ \;
 
\For{$(i=1\rightarrow |\vec{\alpha}|)$}{
    $y_i \leftarrow $ \texttt{gamma\_random($\vec{\alpha_i},1$)}\;
    $\textrm{sum} \leftarrow \textrm{sum} + y_i$\;
}
\For{$(i=1\rightarrow |\vec{\alpha}|)$}{
    $y_i \leftarrow y_i/\textrm{sum}$ \;
}
}
\end{algorithm}
\end{minipage}
}
\caption{Pseudocode for Dirichlet parameter perturbation}
 \label{fig:algo}
\end{figure}

% bookmark  sec.1.4.2  p.6
\subsubsection{Search for the best stochastic matrix with fixed values of $\boldsymbol\alpha$ and $\mathbf{\Theta}$}\label{sec:matrixEstimation}

While holding the Dirichlet and 3-state machine parameters ($\boldsymbol\alpha$ and $\mathbf{\Theta}$) fixed, the current state of the stochastic (base $t=1$) matrix is optimised over all set of sequence-pairs and their alignments in the benchmark $\mathbf{D}$, using a simulated annealing approach.

Starting from an initial temperature parameter of $temp=10000$, we gradually cool down the system using a cooling schedule where the temperature is decreased by a factor of 0.88, after 500 perturbations (described below) of the stochastic matrix at each temperature mark. The process continues until 0.0001 temperature. 

In each perturbation of $\mathbf{M}$, a column %of $\mathbf{M}$ 
of the current matrix
is randomly selected according to its stationary distribution.
%of the current state of matrix $\mathbf{M}$. 
The perturbation of the selected column vector is done by sampling from a Dirichlet distribution with the column vector used as the mean \dirMean{} and with a specified high concentration parameter $\kappa$. The pseudocode of the function \texttt{sample\_dirichlet}$(\vec{\alpha}=\hat{\mu}\kappa)$ has been previously shown in Figure SF1. Initially $\kappa$ is set to $1,000,000$.  As the simulated annealing cools (by a factor of 0.88), the distribution is made even more concentrated by increasing the $\kappa$ by a factor of $\frac{1}{0.88}$, thus making the neighbourhood of sampling tighter and tighter as the simulated annealing cools and the matrix $\mathbf{M}$ converges. 

For each perturbation of $\mathbf{M}\rightarrow\mathbf{\tilde{M}}$, the total lossless encoding length as per Equation 2 (see main text Section 3.2) is recomputed using $\mathbf{\tilde{M}}$. The matrix is accepted or rejected based on the Metropolis criterion: if the encoding length decreases, then the change is accepted with a probability of 1; Otherwise, it is accepted with a probability of ${2^{-\frac{\Delta I}{temp}}}$,
where $\Delta I$ is the difference in the encoding lengths using $\mathbf{\tilde{M}}$ versus $\mathbf{{M}}$ in bits.

% bookmark  p.6  sec.1.5   24/9/2020
%
\subsection{Additional information}
\subsubsection{Computing the expected change of a stochastic matrix} \label{sec:expectedChange}
The expected change of a stochastic matrix modelling amino acid substitutions gives the probability of observing a change in any amino acid state on average. In any column of the matrix  $\mathbf{M}(t)$ indexed by $j$, the diagonal element $\mathbf{M}_{jj}(t)$ represents the probability of observing a \emph{conservation} (i.e. no change) in the corresponding amino acid state, under the substitution model that $\mathbf{M}$ specifies. The weighted average  of these terms (i.e. weighted by the stationary probability $\pi_j$ of each amino acid) over all columns, $\sum_{j=1}^{20}\pi_{j}\mathbf{M}_{jj}(t)$, gives the expected probability of amino acids being conserved under the stochastic model $\mathbf{M}(t)$.
Thus: %, the expected \emph{change} is 
\begin{align*}
    \text{Expected \emph{change}}= 1.0 - \sum_{j=1}^{20}\pi_{j}\mathbf{M}_{jj}(t)
\end{align*}

% bookmark  p.7  sec.1.5.2  24/9/2020
%
\subsubsection{Converting a log-odds scoring matrix into a stochastic matrix} \label{sec:nonMarkovToMarkov}

The approach of establishing a common ground for a fair and consistent comparison between various substitution matrices was made on the basis of converting all non-Markov models into Markov models. Following describes the criteria used for this transformation. 
\begin{enumerate}
    \item  Convert a published, log odds scoring matrix $S$ into its conditional probability matrix form using the following relationship:
        \begin{align} \label{eq:mainScoringMatFormula}
    S_{ij} = c.\log_{2}{\bigg[\frac{\text{\prob{$a_i,a_j$}}}{\text{\prob{$a_i$}\prob{$a_j$}  }} \bigg]}
    = c.\log_{2}{\bigg[\frac{\text{\prob{$a_i$\given$a_j$}}}{\text{\prob{$a_i$}}}\bigg]}
    = c.\log_{2}{\bigg[\frac{\text{\prob{$a_j$\given$a_i$}}}{\text{\prob{$a_j$}}}\bigg]}
\end{align}
where $a_i$ and $a_j$ are the amino acids indexed by $i$ and $j$ respectively; $c$ is the scaling factor to set the unit of information as $\frac{1}{c}$ bits (e.g. $c=2$ implies half bit unit; $c=3$ implies third bit unit). In case the original amino acid null frequencies are absent, we used the 20-nomial model ($\mathbf{P}$) which is optimal under the \texttt{SCOP2} benchmark in order to include it in the fairest way possible. 

\item If the available/derived substitution probability matrix is not reflecting a $0.01$ expected change, assume it as representing some \submat{t} of a Markov model and derive its approximate base matrix $\tilde{\mathbf{M}}(1)$ by finding the $k^{\text{th}}$ matrix root which is the closest to $0.01$ expected change, which is then normalized.

%\item Scale the resultant base matrix with a current expected change \texttt{curr} to possess a $0.01$ expected change, under the assumption that evolutionary distance $t$ is linear to the expected change within a small interval $t \pm \delta t$.
%\begin{align*}
%    \text{\submat{1}} = \tilde{\mathbf{M}}(1)^{\frac{0.01}{\texttt{curr}}} 
%\end{align*}
\end{enumerate}

% bookmark  p.8  sec.2  24/9/2020
%
\section{Supplementary Results}

\subsection{Distribution sequence-identity in the benchmarks}
The figure below gives the sequence-identity distribution of alignments in each of the six benchmarks we considered. This accompanies the discussion of benchmarks in results section 2.1 of the main text.
\begin{figure}[!h] 
    \centering
    \includegraphics[width=\textwidth]{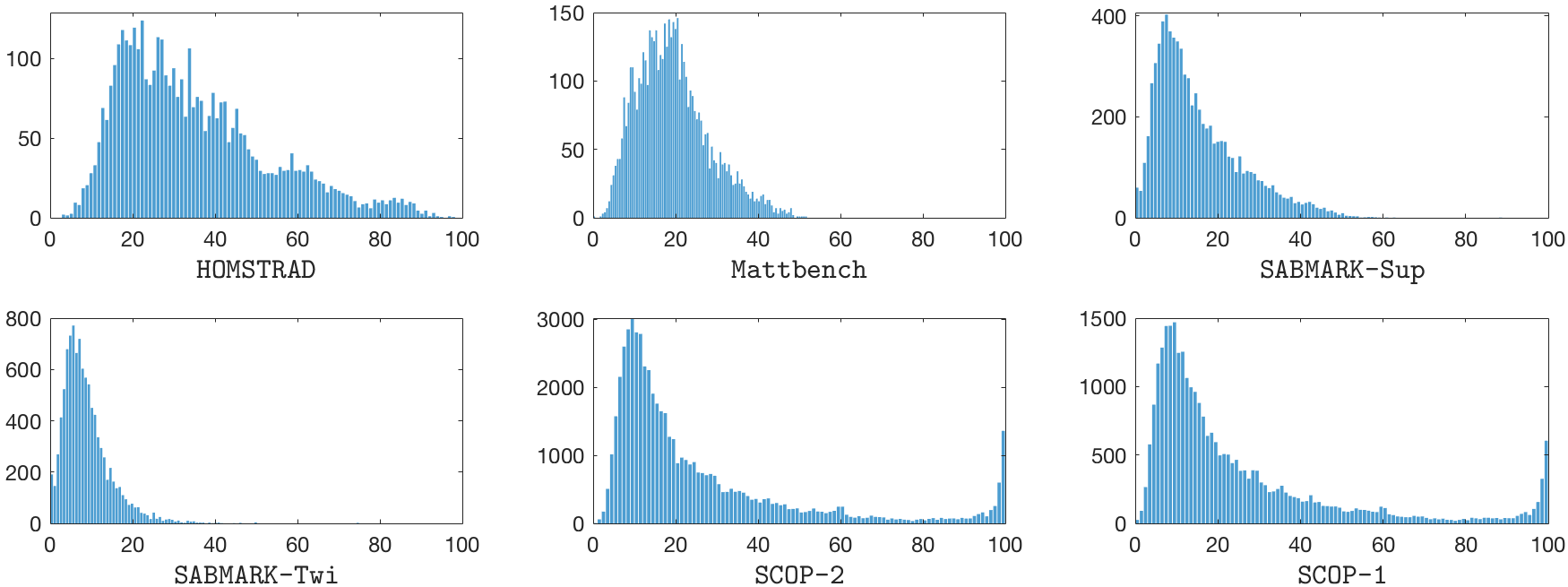}
    \caption{Sequence identity percentage distributions (histograms) across the benchmarks}
    \label{fig:benchmarkHist} 
\end{figure}

% bookmark  sec.2.2  p.8  ------------------------------------------------------------------------------
\subsection{Extended Analysis accompanying results in main text (Section 2.2)}
\label{sec:sm-ExtendedAnalysis}

\paragraph{Shannon information content across matrices and benchmarks by varying $\mathbf{P}$.} 

Table 2 of the main text (see results section 2.2)  examined the measure of Shannon information content by varying the substitution matrix model, while all other parameters (including the 20-nomial probabilities $\mathbf{P}$) were optimised for the respective matrices and benchmarks. 

As an extended analysis, we explore two other alternatives for  $\mathbf{P}$ to study their effect on the rank order, although they turned out to be suboptimal compared to the optimal ones reported in the main text, Table 2. These alternatives are:
\begin{enumerate} 
\item inferring a 20-nomial from an independent-source -- here, we use the entire set of non-redundant protein sequences in the UniProt database~\cite{uniprot2017};
\item using the stationary distribution of the stochastic matrix $\mathbf{M}$ (i.e. derived from the eigenvector of $\mathbf{M}$ associated with the eigenvalue of $\lambda=1$).
\end{enumerate}

\myTable{tab:totalResults2} shows the Shannon information content across the matrices and benchmarks, for the first choice enumerated above. We observe that, although the encoding lengths are longer (worser), the effect on the rank order is minimal. The order (based on \texttt{ranksums}) observed in Table 2 of the main text are generally preserved, barring \texttt{LG} and \texttt{MIQS} where their ranks over \texttt{SCOP2} have swapped over. In comparison,  \myTable{tab:totalResults3} shows the same encoding lengths using the second choice enumerated above, and there are some changes to the rank order, although MML$_\texttt{SCOP2}$ matrix still performs the best across all benchmarks with the lowest \texttt{ranksum=16}. 

%Overall, $\mathbf{P}$ inferred optimally on the indel regions of a benchmark (see Table 2 of the main text) gives the best results (i.e. the shortest encoding lengths).
%Next best choice of $\mathbf{P}$ is to set it to %stationary distribution of the matrix (see \myTable{tab:totalResults3}), whereas the $\mathbf{P}$ inferred on UniProt gives the longest encoding lengths (see \myTable{tab:totalResults2}).

\begin{table}[!h]
    \caption{\small Shannon information content to losslessly encode each benchmark with varying substitution matrices along with their ranks (reported inside brackets), when using the UniProt based MML estimates as the amino acid 20-nomial distribution $\mathbf{P}$.}
\scriptsize
    \centering
    \begin{tabular}{|l|c|c|c|c|c|c|}
    \hline
    \textbf{Matrix} &  \texttt{\textbf{HOMSTRAD}} &  \texttt{\textbf{Mattbench}} &  \texttt{\textbf{SABMARK-Sup}} & \texttt{\textbf{SABMARK-Twi}}  &  \texttt{\textbf{SCOP-1}}  & \texttt{\textbf{SCOP-2}} \\ 
    \hline \hline 
    \multicolumn{7}{|c|}{Shannon information using existing substitution models}\\
    \hline \hline
\texttt{PAM} (1978)   		& 11547347.32 (15) &  9155559.27 (15) &  23628852.13 (15) &  11336736.12 (15) &  85026098.24 (15) &  82893663.55 (15) \\
\texttt{JTT} (1992)    		& 11496994.08 (14) &  9084491.06 (13) &  23505598.27 (13) &  11278424.47 (13) &  84454677.89 (13) &  82354250.08 (13) \\
\texttt{BLOSUM} (1992)   	& 11453343.72 (10) &  9049472.26 (08) &  23428674.77 (07) &  11254553.75 (06) &  84275402.19 (11) &  82130897.40 (10) \\
\texttt{JO} (1993)           & 11492309.39 (13) &  9130688.40 (14) &  23556128.16 (14) &  11316566.45 (14) &  84668169.16 (14) &  82541280.05 (14) \\
\texttt{WAG} (2001)    		& 11434976.91 (05) &  9065145.32 (12) &  23454783.64 (11) &  11269752.08 (12) &  84242325.21 (09) &  82131872.38 (11) \\
\texttt{VTML} (2002)   		& 11439289.05 (07) &  9048325.84 (07) &  23432271.63 (08) &  11257134.26 (08) &  84176599.86 (07) &  82061020.74 (07) \\
\texttt{LG}  (2008)  	 	    & 11480054.48 (12) &  9061463.01 (11) &  23466479.90 (12) &  11261899.34 (09) &  84356348.28 (12) &  82226007.09 (12) \\
\texttt{MIQS} (2013)  		& 11438006.26 (06) &  9052903.26 (10) &  23440009.45 (10) &  11262833.45 (10) &  84177434.21 (08) &  82063425.69 (08) \\
\texttt{PFASUM} (2017)   	& 11428679.05 (02) &  9052221.77 (09) &  23433841.03 (09) &  11263082.39 (11) &  84141210.67 (04) &  82038431.69 (06) \\

    \hline \hline 
    \multicolumn{7}{|c|}{Shannon information using MML inferred substitution models}\\
    \hline \hline
    
MML$_\text{\texttt{HOMSTRAD}}$  & 11421395.62 (01) &  9047739.99 (05) &  23419917.83 (06) &  11256695.07 (07) &  84126993.63 (03) &  82009294.02 (03) \\
MML$_\text{\texttt{MATTBENCH}}$  & 11442135.03 (09) &  9038304.82 (01) &  23409982.48 (03) &  11243729.20 (03) &  84151618.78 (05) &  82022514.41 (04) \\
MML$_\text{\texttt{SABMARK-Sup}}$   & 11439926.80 (08) &  9043738.29 (04) &  23400792.07 (01) &  11238762.79 (02) &  84168584.28 (06) &  82024870.38 (05) \\
MML$_\text{\texttt{SABMARK-Twi}}$ & 11458572.58 (11) &  9048142.90 (06) &  23410821.13 (05) &  11237871.03 (01) &  84256392.91 (10) &  82098025.94 (09) \\
MML$_\text{\texttt{SCOP1}}$  & 11429086.52 (03) &  9042105.23 (03) &  23410002.48 (04) &  11247805.78 (05) &  84097487.36 (01) &  81984099.12 (02)  \\
MML$_\text{\texttt{SCOP2}}$ & 11429516.19 (04) &  9041089.95 (02) &  23404593.45 (02) &  11244715.72 (04) &  84100227.75 (02) &  81976372.74 (01)  \\
\hline
    \end{tabular}
    \label{tab:totalResults2}
\end{table}

\begin{table}[!h]
        \caption{\small Shannon information content to losslessly encode each benchmark with varying substitution matrices along with their ranks (reported inside brackets), when using the matrix stationary distribution as the amino acid 20-nomial distribution $\mathbf{P}$.}
    
\scriptsize
    \centering
    \begin{tabular}{|l|c|c|c|c|c|c|}
    \hline
    \textbf{Matrix} &  \texttt{\textbf{HOMSTRAD}} &  \texttt{\textbf{Mattbench}} &  \texttt{\textbf{SABMARK-Sup}} & \texttt{\textbf{SABMARK-Twi}}  &  \texttt{\textbf{SCOP-1}}  & \texttt{\textbf{SCOP-2}} \\ 
    \hline \hline 
    \multicolumn{7}{|c|}{Shannon information using existing substitution models}\\
    \hline \hline
    
\texttt{PAM} (1978)  		& 11547416.21 (15) &  9164067.18 (15) &  23657379.78 (15) &  11355479.09 (15) &  85072048.36 (15) &  82949452.90 (15) \\
\texttt{JTT} (1992)   		& 11496340.10 (14) &  9084284.48 (13) &  23499066.75 (13) &  11273707.07 (11) &  84446694.25 (13) &  82346379.77 (13) \\
\texttt{BLOSUM} (1992)   	& 11459479.32 (11) &  9060438.92 (10) &  23466561.57 (11) &  11275166.62 (13) &  84354008.80 (11) &  82232019.26 (11)  \\
\texttt{JO} (1993) 		    & 11490030.65 (13) &  9139874.62 (14) &  23571493.08 (14) &  11328875.88 (14) &  84680102.91 (14) &  82560968.37 (14) \\
\texttt{WAG} (2001) 		& 11432008.93 (07) &  9065038.48 (11) &  23442803.68 (10) &  11262564.43 (10) &  84218617.83 (10) &  82107643.24 (10) \\
\texttt{VTML} (2002) 		& 11439908.46 (10) &  9049441.74 (07) &  23432043.81 (09) &  11255766.35 (09) &  84180147.95 (09) &  82070311.94 (09) \\
\texttt{LG}  (2008) 	 	    & 11485493.34 (12) &  9065759.17 (12) &  23491875.49 (12) &  11274788.00 (12) &  84399024.43 (12) &  82286691.96 (12) \\
\texttt{MIQS} (2013)  		& 11435028.94 (08) &  9051333.90 (09) &  23421817.75 (08) &  11252372.16 (08) &  84147813.60 (08) &  82031386.28 (08) \\
\texttt{PFASUM} (2017)   	& 11423630.07 (02) &  9048568.06 (06) &  23406776.08 (05) &  11248125.05 (06) &  84095385.87 (04) &  81984520.86 (04)  \\

    \hline \hline 
    \multicolumn{7}{|c|}{Shannon information using MML inferred substitution models}\\
    \hline \hline

MML$_\text{\texttt{HOMSTRAD}}$   & 11418789.53 (01) &  9050681.63 (08) &  23408123.19 (06) &  11250120.55 (07)  &  84113701.63 (06) &  82002419.36 (06) \\
MML$_\text{\texttt{MATTBENCH}}$  & 11436960.04 (09) &  9040249.80 (01) &  23416280.31 (07) &  11248122.03 (05)  &  84130366.58 (07) &  82027458.07 (07) \\
MML$_\text{\texttt{SABMARK-Sup}}$  & 11423841.26 (03) &  9044793.01 (05) &  23401281.01 (01) &  11244464.40 (02)  &  84092356.97 (03) &  81982211.04 (03) \\
MML$_\text{\texttt{SABMARK-Twi}}$& 11427701.22 (06) &  9043437.94 (02) &  23404974.85 (04) &  11243567.19 (01)  &  84099460.19 (05) &  81990981.18 (05) \\
MML$_\text{\texttt{SCOP1}}$  & 11424651.38 (05) &  9043895.49 (04) &  23403924.44 (03) &  11244987.34 (03)  &  84075928.68 (01) &  81970279.92 (02) \\
MML$_\text{\texttt{SCOP2}}$  & 11423906.32 (04) &  9043878.06 (03) &  23403244.25 (02) &  11245052.04 (04)  &  84077082.89 (02) &  81968197.28 (01)  \\
\hline
    \end{tabular}
    \label{tab:totalResults3}
\end{table}

% bookmark  p.9  sec.2.2.2  24/9/2020
%

% bookmark  p.10  sec.2.3  24/9/2020
%
\section{Additional information}
%\subsection{Analysis of encoding lengths when converting and using log-odds substitution scoring matrices from BLOSUM and PFASUM series.}
\subsection{Analysis of encoding lengths using BLOSUM and PFASUM series of matrices}
%Supplementary section S1.5.2 presents the method of converting any log-odds scoring matrix into a stochastic matrix. 
Here we analyse how all matrices in the BLOSUM series (most popularly used substitution matrix series) and PFASUM series (the best-performing, BLOSUM-like series among the existing set of substitution matrices compared in this work) perform in compressing the \texttt{SCOP2} benchmark. 

%the %effect on the
%encoding lengths (i.e. Shannon information content) of all
%when converting log-odds 
%matrices in the BLOSUM series (most popularly used non-Markov substitution model) and PFASUM series (the best-performing, BLOSUM-like series among the existing set of substitution matrices compared here in this work) %into 
%under their respective stochastic form.% and using them to compress the same benchmark.

\begin{figure}[!h]
    \centering
    \includegraphics[scale=0.2]{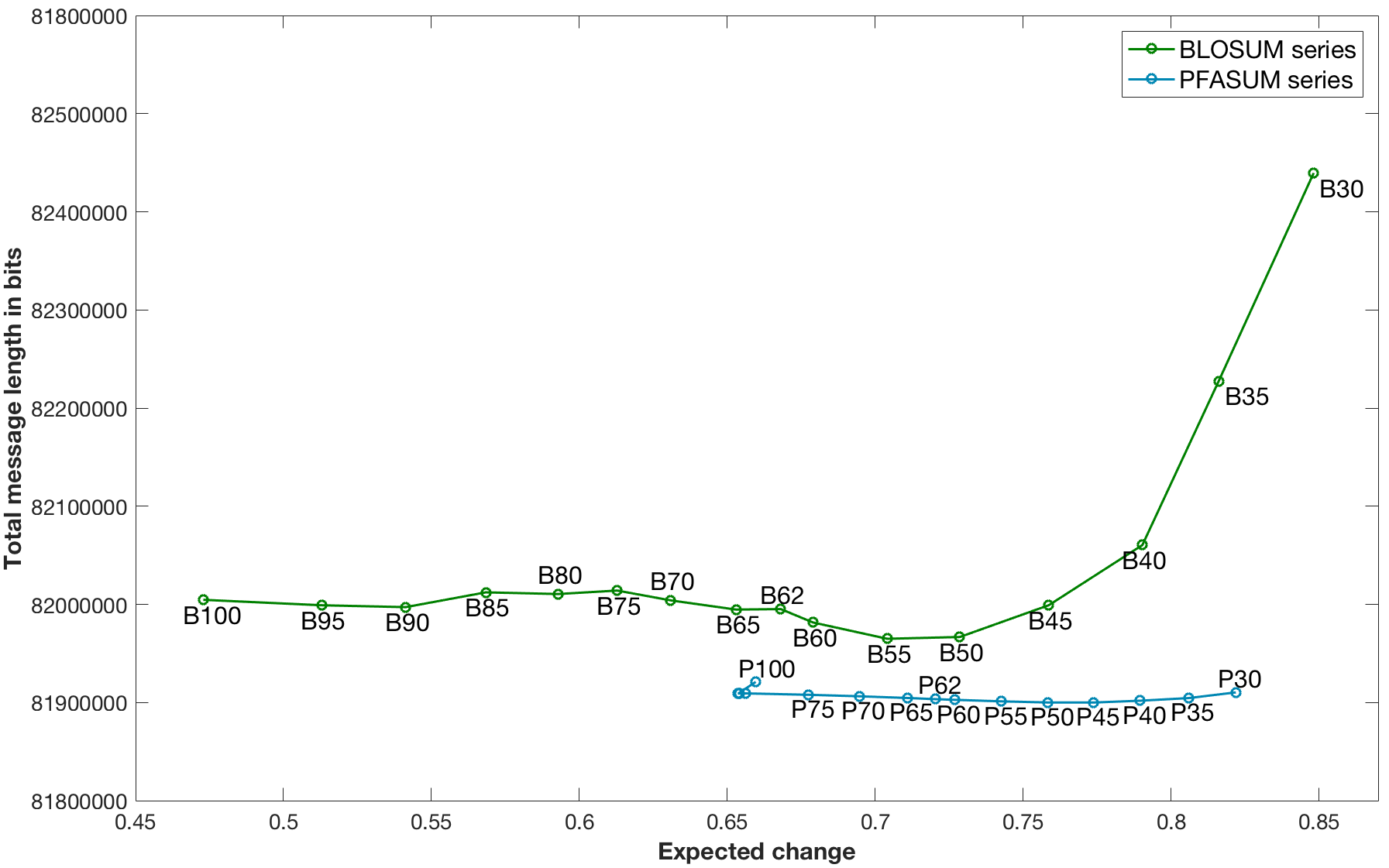}
    \caption{\small Lossless encoding lengths resulted from using matrices in the BLOSUM and PFASUM series to compress \texttt{SCOP2} benchmark, plotted against their original expected change of amino acids.}
    \label{fig:BlosumPfasum}
\end{figure}

In Table 2 of the main text, the rows reporting the encoding lengths for BLOSUM and PFASUM were by using the converted forms of BLOSUM62 and PFASUM60 matrices respectively. (These were chosen as they are the reported general-purpose log-odds scoring matrices in their respective series.) Table \ref{tab:BlosumPfasum} extends this analysis to other possible matrices in their series, all compressing the same \texttt{SCOP2} benchmark. The accompanying Figure \ref{fig:BlosumPfasum} illustrates the total encoding message lengths (i.e. Shannon information content) reported in Table \ref{tab:BlosumPfasum} as a function of the expected amino acid change implicit in %various scoring matrices in the two series.
their original scoring matrices.

Consistent to the picture that emerged when comparing existing matrices, we observe (see Figure \ref{fig:BlosumPfasum}) that the PFASUM series performs better than BLOSUM series. Across the matrices in PFASUM, we observe a flat trend of encoding lengths suggesting their internal consistency, and that any of them are equally good to derive a stochastic matrix from.  On the other hand, BLOSUM series, with some  variations, has more or less a flat trend when using matrices converted into a stochastic form within the range [45,100], whereas matrices in the range [30,40] give substantially worse (larger) message lengths in comparison. This suggests that  BLOSUM[30-40] series of matrices do not generalise well to explain the range of relationships in \texttt{SCOP2} benchmark. 

\begin{table}[!h]
    \caption{\small Lossless encoding lengths (in bits) resulted from using various substitution matrices from the BLOSUM and PFASUM series, to compress the \texttt{SCOP2} benchmark.} 
\scriptsize
    \centering
    \begin{tabular}{|l|c|c||}
    \hline 

\parbox{1.4cm}{\centering \textbf{BLOSUM series} }
& \parbox{1.4cm}{\centering \textbf{Original expected change}  } & \parbox{1.4cm}{\centering \textbf{Total message length}}  \\ 
    \hline 
    \hline 
BLOSUM30 &      0.8482  & 82439452.76  \\
BLOSUM35 &      0.8161  & 82227426.03   \\
BLOSUM40 &      0.7903  & 82060377.07   \\
BLOSUM45 &      0.7587  & 81999170.77    \\
BLOSUM50 &      0.7286  & 81966740.66   \\
BLOSUM55 &      0.7041  & 81964908.57   \\
BLOSUM60 &      0.6790  & 81981508.57   \\ 
BLOSUM62 &      0.6681  & 81995179.31    \\
BLOSUM65 &      0.6530  & 81994670.34   \\
BLOSUM70 &      0.6309  & 82004034.57   \\
BLOSUM75 &      0.6128  & 82014056.31   \\
BLOSUM80 &      0.5928  & 82010530.24  \\
BLOSUM85 &      0.5684  & 82012220.79  \\
BLOSUM90 &      0.5413  & 81997158.82  \\
BLOSUM95 &      0.5129  & 81999074.38   \\
BLOSUM100&      0.4729  & 82004612.48   \\
\hline 
    \end{tabular}
    \quad \quad
     \begin{tabular}{|l|c|c||}
    \hline 
\parbox{1.4cm}{\centering \textbf{PFASUM series} } &  \parbox{1.4cm}{\centering \textbf{Original expected change}  }     &\parbox{1.4cm}{\centering \textbf{Total message length}  }  \\ 
    \hline 
    \hline 
PFASUM30  & 0.8219     & 81910293.61  \\
PFASUM35  & 0.8059     & 81904481.06  \\
PFASUM40  & 0.7896     & 81901816.25  \\
PFASUM45  & 0.7739     & 81899988.62  \\
PFASUM50  & 0.7584     & 81900015.41  \\
PFASUM55  & 0.7427     & 81901163.14  \\
PFASUM60  & 0.7268     & 81902713.60  \\ 
PFASUM62  & 0.7204     & 81903374.10  \\
PFASUM65  & 0.7108     & 81904599.34  \\
PFASUM70  & 0.6946     & 81906335.59  \\
PFASUM75  & 0.6774     & 81907860.48  \\
PFASUM80  & 0.6562     & 81909272.54  \\
PFASUM85  & 0.6540     & 81909221.82  \\
PFASUM90  & 0.6538     & 81909091.84  \\
PFASUM95  & 0.6537     & 81909179.07  \\
PFASUM100 & 0.6596     & 81920990.78  \\
\hline 
    \end{tabular}
    \label{tab:BlosumPfasum} 
\end{table}

\subsection{Kullback-Leibler (KL) divergence between matrices}
\label{sec:kldivmat}
We now undertake an analysis to quantitatively highlight the similarities and differences between ten stochastic matrices (the nine existing matrices and MMLSUM (i.e. MML$_{\texttt{SCOP2}}$)) independent of any benchmarks. We emphasise that this analysis does \emph{not} provide any statement on the relative performance of various matrices (as was carried out in the main text and previous section), but will assist in gauging their relative concordance. 
 
This analysis uses the notion of Kullback-Leibler (KL) divergence~\cite{kullback1951information} computed between any two substitution matrices. KL-divergence estimates the measure of \emph{relative} Shannon entropy between two probability distributions.  For any two matrices $X$ and $Y$ (considered in their stochastic form), their KL-divergence is measured in two different ways as follows. 
\begin{eqnarray*}
	\text{KL divergence (over joint probabilities)} &=& \sum_{i=1}^{20} \sum_{j=1}^{20}  X_{i,j}\log{\left(\frac{X_{i,j}}{Y_{i,j}}\right)}  \\
		\text{KL divergence (over conditional probabilities)} 
		&=& \sum_{i=1}^{20} \sum_{j=1}^{20} X_{i,j} \log{\left(\frac{X_{i|j}}{Y_{i|j}}\right)}
\end{eqnarray*}
where $X_{i,j}$ denotes the joint probability implied by the matrix $X$ for the pair of amino acids indexed by $i$ and $j$ (similarly, $Y_{i,j}$), $X_{i|j}$ denotes the conditional probability implied by the matrix $X$ of the amino acids indexed by $i$ given (i.e. interchanging/substituting) an amino acid indexed by $j$ (similarly, $Y_{i|j}$). Note, by default, the stochastic matrix is in its conditional form containing  conditional probability terms between amino acids. The joint probability terms can be computed as a product of its conditional probability times its stationary probability.

\begin{figure}[!t]
    \centering
       \includegraphics[scale=0.25]{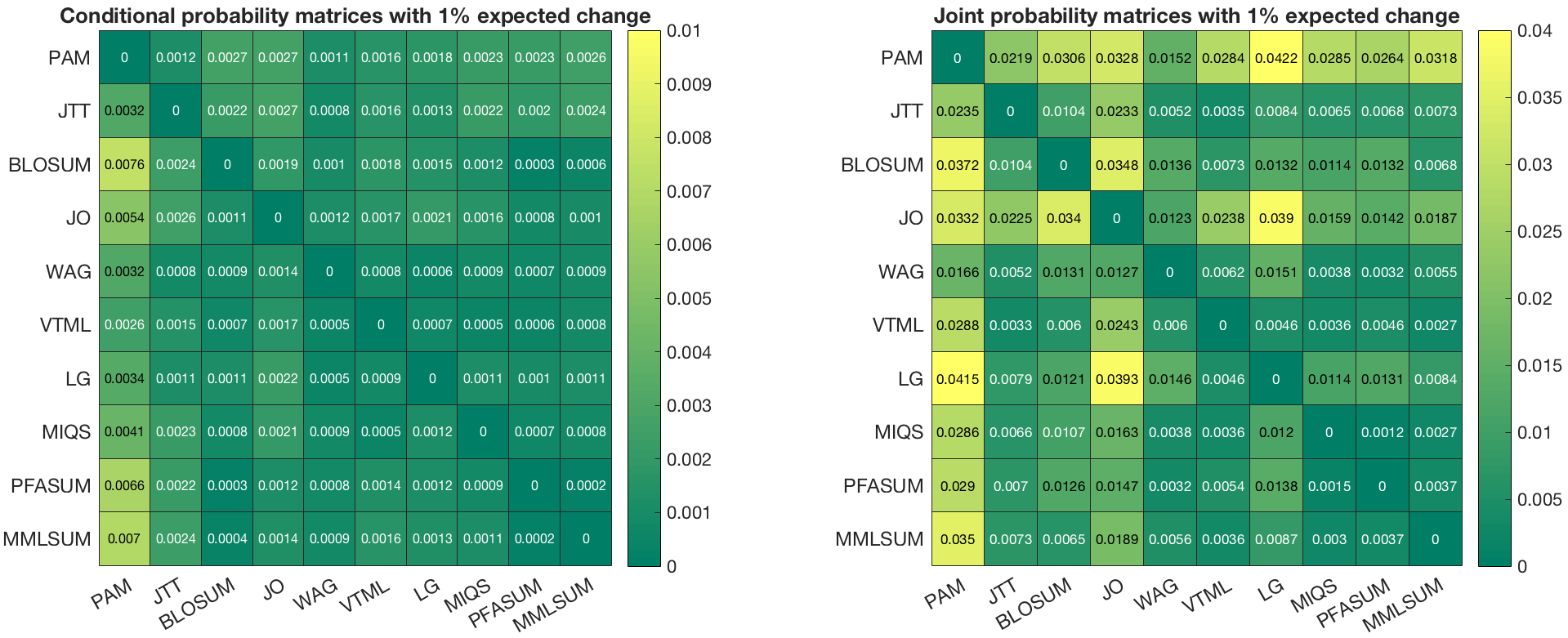}
\caption{\small KL divergence based distance matrices (in the form of heat maps)
between 10 matrices (9 existing and 1 inferred using MML). Left column is the computation of KL-divergence over conditional probabilities of amino acid interchanges coming under the base matrix of $1\%$ expected change. Right column is the computation of KL-divergence over the corresponding joint probabilities of interchanges. %The three rows compare between equivalent states of matrices when their expected amino acid change is $1\%, 50\%$ and $80\%$ respectively.
 \label{fig:klmat}}
\end{figure}

%\ref{fig:klmat} gives the KL-divergence across the aforementioned conditional and joint measures across matrices. To highlight the relative-change in the concordance between matrices as the matrices evolve with time, \myFig{fig:klmat} shows KL-divergence  computed between matrices at equivalent time points. These time points are equivalent in the sense that the estimated expected change in amino acids are all same. 

Figure \ref{fig:klmat} shows the KL-divergence -- for conditional probabilities in left column; for joint probabilities in right column -- between all matrices when they show an expected amino acid change of 1\%. (Note, KL-divergence is not a metric, so the resultant table is not symmetric.)  
%We notice that as the expected change of amino acids increases, the differences between matrices become more apparent.  
%\texttt{PAM} and \texttt{JO} are radically different from other matrices, and this correlates with the their poor performance in terms of their \texttt{ranksums} discussed in the main text. 
%\arun{Beyond saying Fig. 1. is pretty, I am struggling to put a 'spin' on this that supports its presence in the main text. Have failed in my attempts. Also, those clustering trees on top and leftside of Fig. 1. are NOT same? What do they mean??? I recommend, we remove them both, get the table ordered in the original order of matrices shown in Table 2, as it may help keep the story simple::}

% bookmark  sec.2.4  p.11  24/9/2020
%
\subsection{Stochastic matrix convergence to steady-state probabilities}

\begin{figure*}[!h]
\centering 
\includegraphics[width=\textwidth]{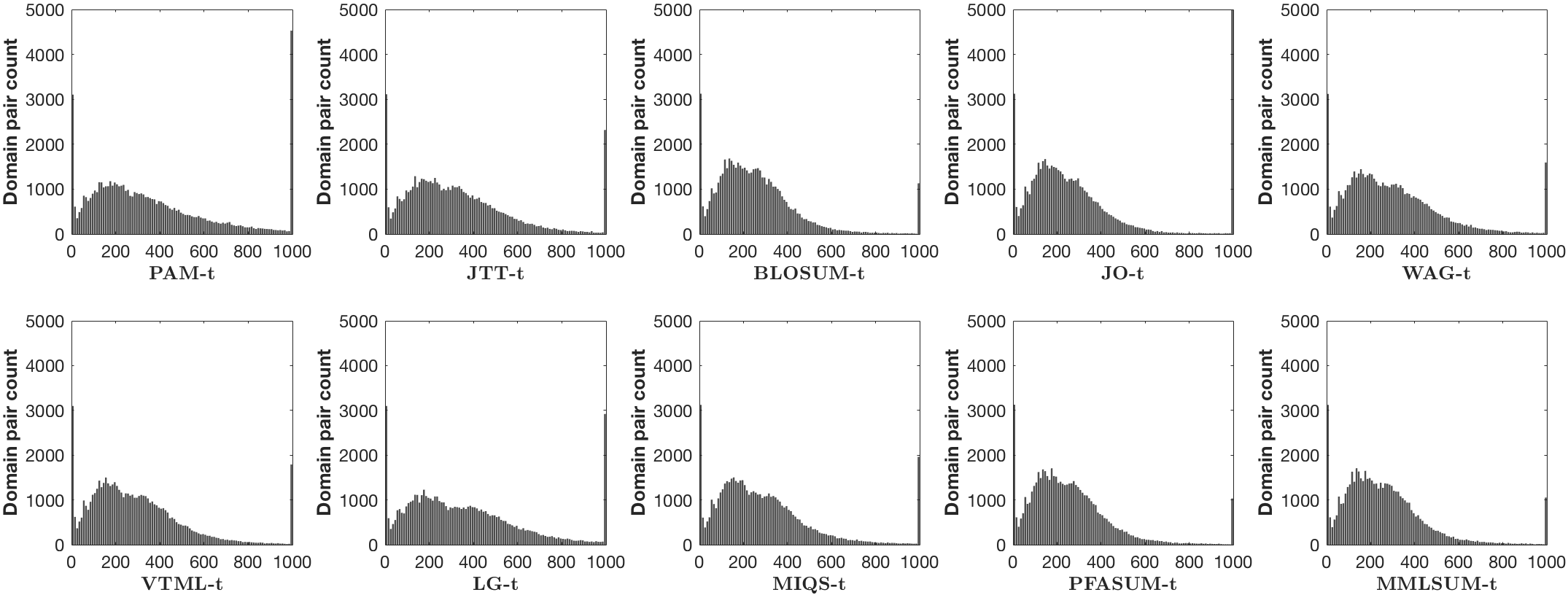}
\caption{\small Histograms of the sequence-divergence parameter $t$ under each substitution model, inferred for the $59,092$ protein domain pairs in the \texttt{SCOP2} benchmark}
\label{fig:evolDist1}
\end{figure*}

The optimal evolutionary distance $t$ between two amino acid sequences may vary under different substitution models. Figure \ref{fig:evolDist1} illustrates the histograms of $t$ resultant under different stochastic matrices when
encoding the \texttt{SCOP2} benchmark. 

\begin{figure*}[!t] 
	\includegraphics[width=\textwidth]{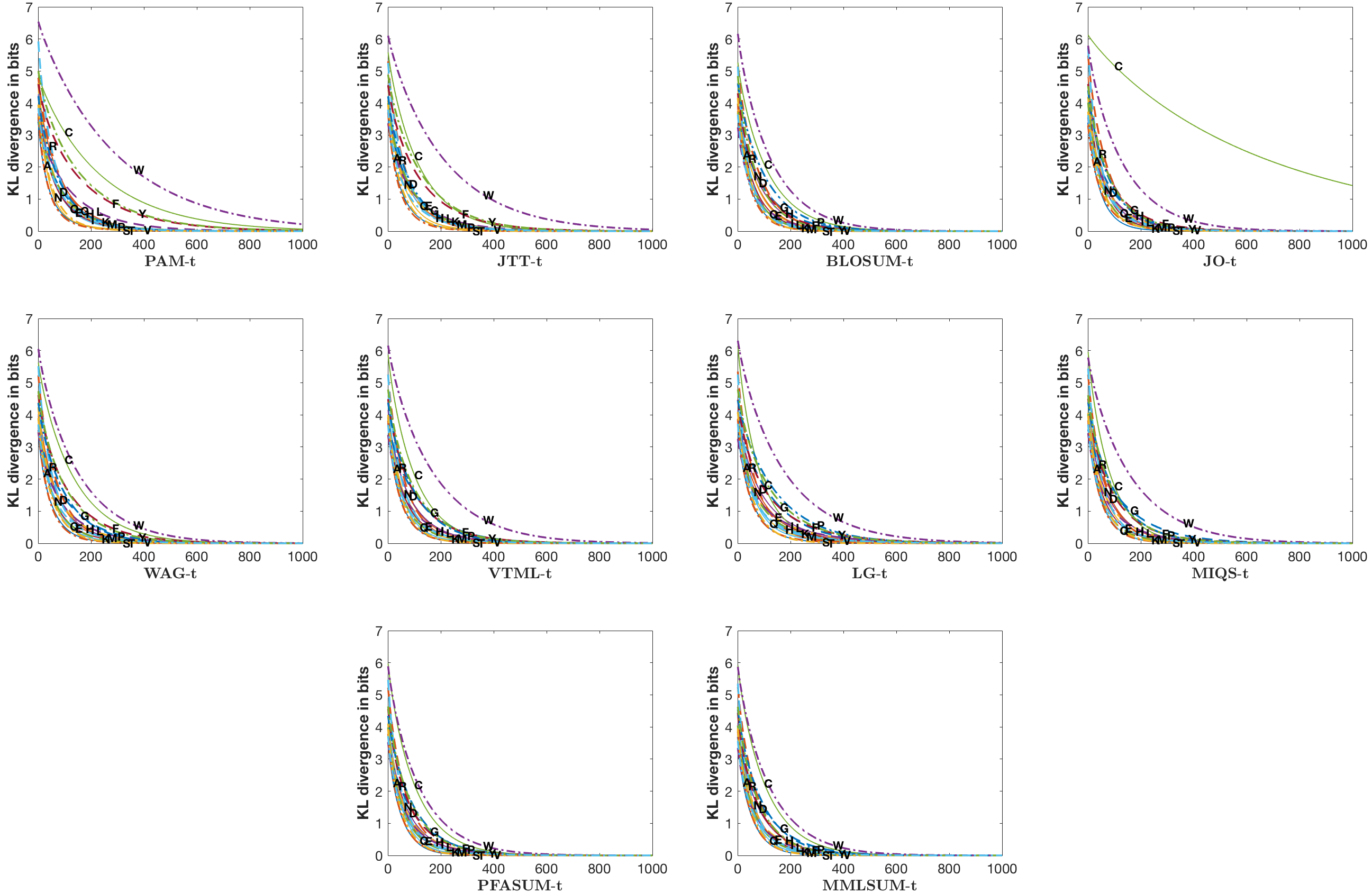}
        \caption{The KL-divergence measuring the convergence of each column vector of various matrices to their respective stationary probabilities.}
\label{fig:kldivplots}
\end{figure*}

Fore each matrix, the convergence of amino acids to their respective stationary probabilities is shown in Figure \ref{fig:kldivplots}). 
Here, the KL divergence was measured between each amino acid column of a stochastic (conditional probability) matrix and the matrix's stationary distribution. Once an amino acid $x$ column reaches the steady state, the matrix has a probability of interchange of $x$ with any amino acid, that is indistinguishable from its stationary probability.  

%As observed in Figure \ref{fig:kldivplots},  only \texttt{W}  is an outlier common across all matrices, taking more time to converge compared to other amino acids. \texttt{C} is an outlier for most matrices, except LG and MIQS models.  PAM  has four amino acids as an outlier to the general trend: $\{$\texttt{W},\texttt{C}, \texttt{F}, \texttt{Y}$\}$.
%The smaller spectral gap in JO must be mainly due to this outlier column. 
%Figure \ref{fig:kldivplotmmlsum} zooms into the KL divergence plot of MMLSUM stochastic matrix. All curves roughly tend to 0 around $t=600$. 

\ignore{
\begin{figure}
    \centering
    \includegraphics[width=\textwidth]{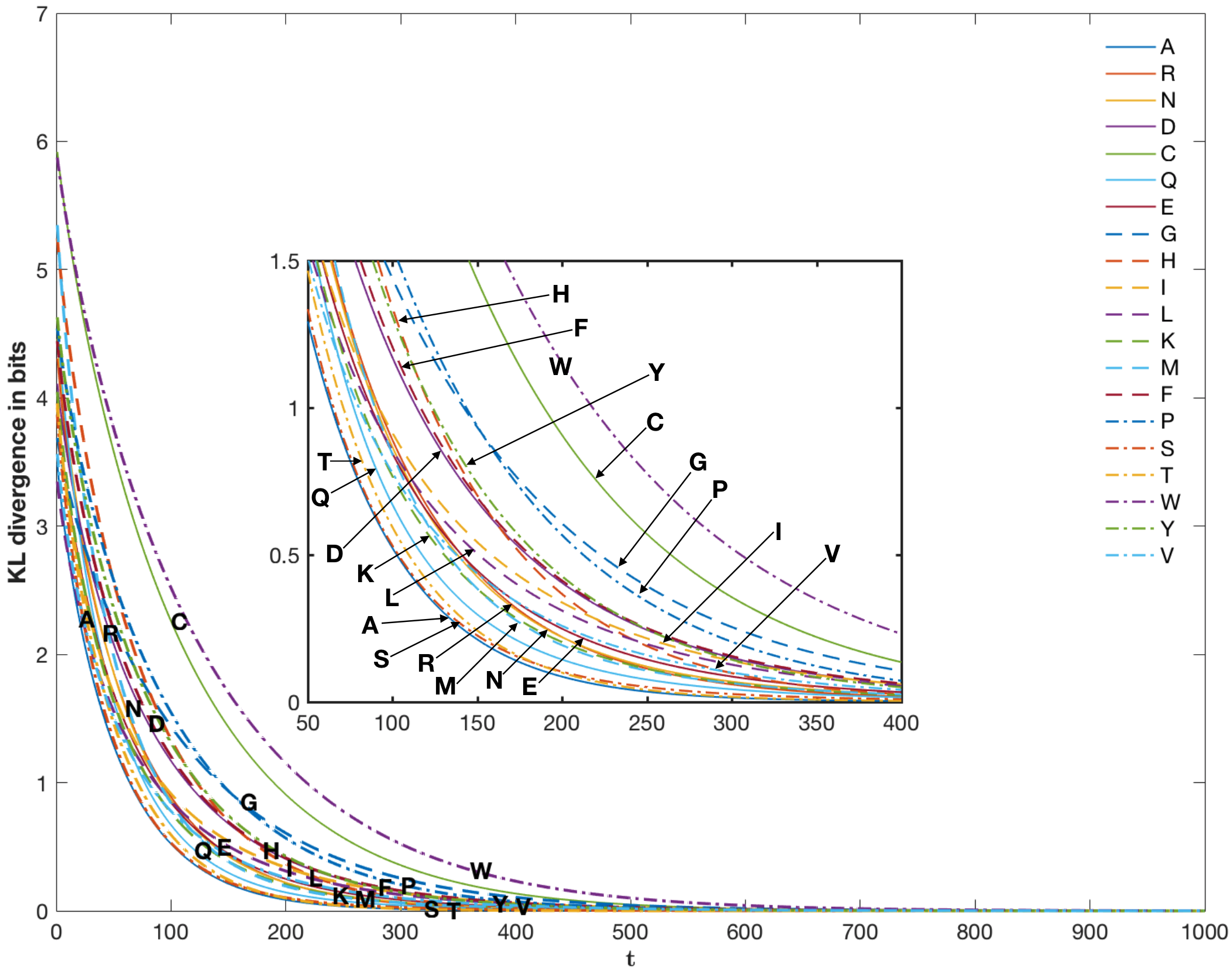}
    \caption{How each column vector of the MMLSUM conditional probability matrix reaches its stationary distribution as the evolutionary distance $t$ increases, measured in terms of the KL divergence between the two $\mathbb{L}1$-normalised probability vectors. Note that by $t=600$, all columns are almost at equilibrium}
    \label{fig:kldivplotmmlsum}
\end{figure}

%The difference in clock speed also relates to how the expected change of each model differs as a function of $t$, shown in Figure \ref{fig:expectedchangeall}. 
%Separately, the Relative entropy of a substitution matrix provides a measure of how well the matrix differentiates relatedness from un-relatedness at a given \evolDist{}, based on its corresponding joint probability distribution and stationary distribution.

\begin{figure*}[!b]
\includegraphics[width=\textwidth]{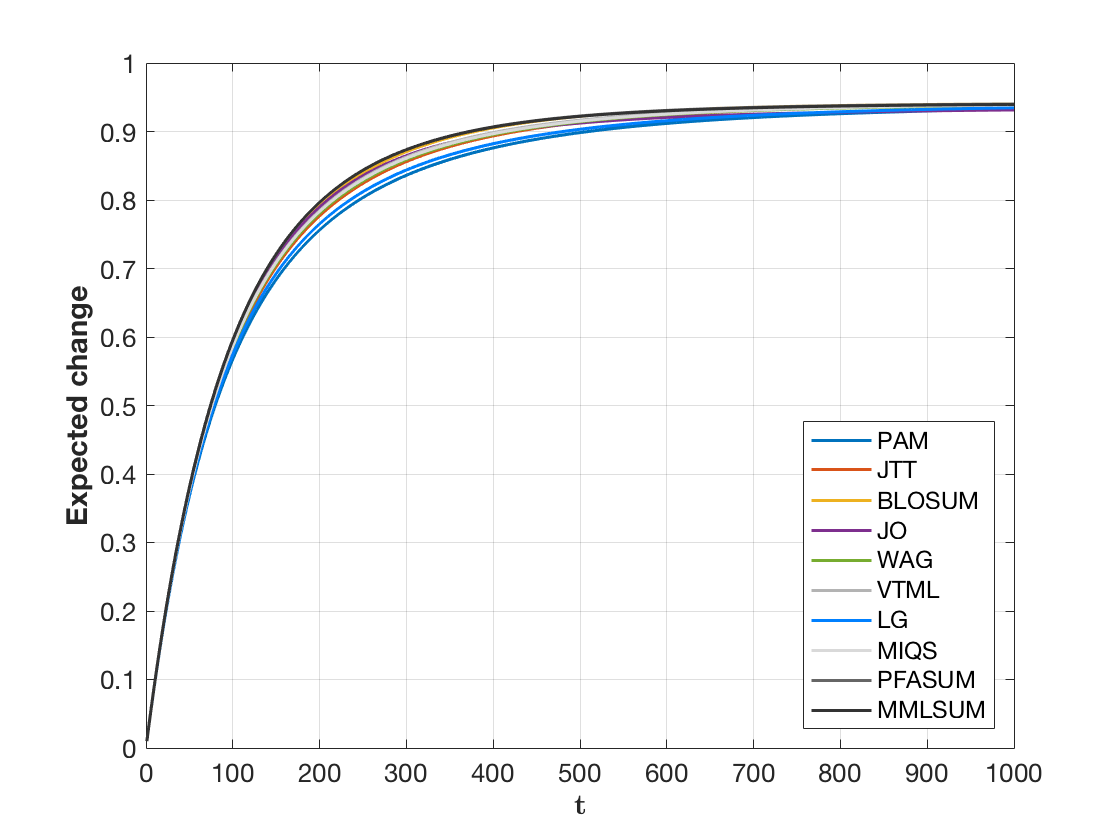}
\caption{How the expected change %and (b) relative entropy 
of each Markov model change as a function of evolutionary distance $t$}
\label{fig:expectedchangeall}
\end{figure*}

%\begin{figure}[!tb]
 %   \centering
 %   \includegraphics[scale=0.2]{Figures/EigTrend.png}
 %   \caption{How the second largest eigenvalue of the stochastic matrix decreases with evolutionary distance $n$}
 %   \label{fig:eigTrend}
%\end{figure}

}

\clearpage

\bibliographystyle{abbrvnat}
\bibliography{references}